\documentclass[conference]{IEEEtran}
\usepackage{times}
\usepackage{graphicx}
\usepackage{tikz}
\usepackage{tabularx}
\usepackage{booktabs}
\usepackage{amssymb}
\usepackage{bm}
\usepackage{amsmath}
\usepackage{threeparttable}
\usepackage{multirow}
\usepackage{cuted}
\usepackage{capt-of}
\usepackage{subcaption}

\usepackage[numbers]{natbib}
\usepackage{multicol}
\usepackage[bookmarks=true]{hyperref}
\graphicspath{{figures/}}

\definecolor{mygreen2}{RGB}{0 180 0}
\hypersetup{
    colorlinks=true,
    linkcolor=magenta,
    urlcolor=magenta,
    citecolor=mygreen2,
}

\begin{document}

\title{Toward Reliable Sim-to-Real Predictability for MoE-based Robust Quadrupedal Locomotion}

\author{
    Tianyang Wu$^{1}$, Hanwei Guo$^{1}$, Yuhang Wang$^{1}$, Junshu Yang$^1$, Xinyang Sui$^1$, Jiayi Xie$^1$, \\
    Xingyu Chen$^{1*}$, Zeyang Liu$^{1*}$, Xuguang Lan$^{1*}$ \\
    \small $^1$National Key Laboratory of Human-Machine Hybrid Augmented Intelligence,\\Institute of Artificial Intelligence and Robotics, Xi'an Jiaotong University,
    $^*$Corresponding Authors \\
    Page: \href{https://robogauge.github.io/complete/}{\texttt{https://robogauge.github.io/complete/}} \quad Code: \href{https://github.com/wty-yy/go2_rl_gym}{\texttt{Train}}, \href{https://github.com/wty-yy/RoboGauge}{\texttt{Evaluate}}, \href{https://github.com/wty-yy/unitree_cpp_deploy}{\texttt{Deploy}}
}



%

\makeatletter
\let\@oldmaketitle\@maketitle
    \renewcommand{\@maketitle}{\@oldmaketitle
    \centering
    \includegraphics[width=\linewidth]{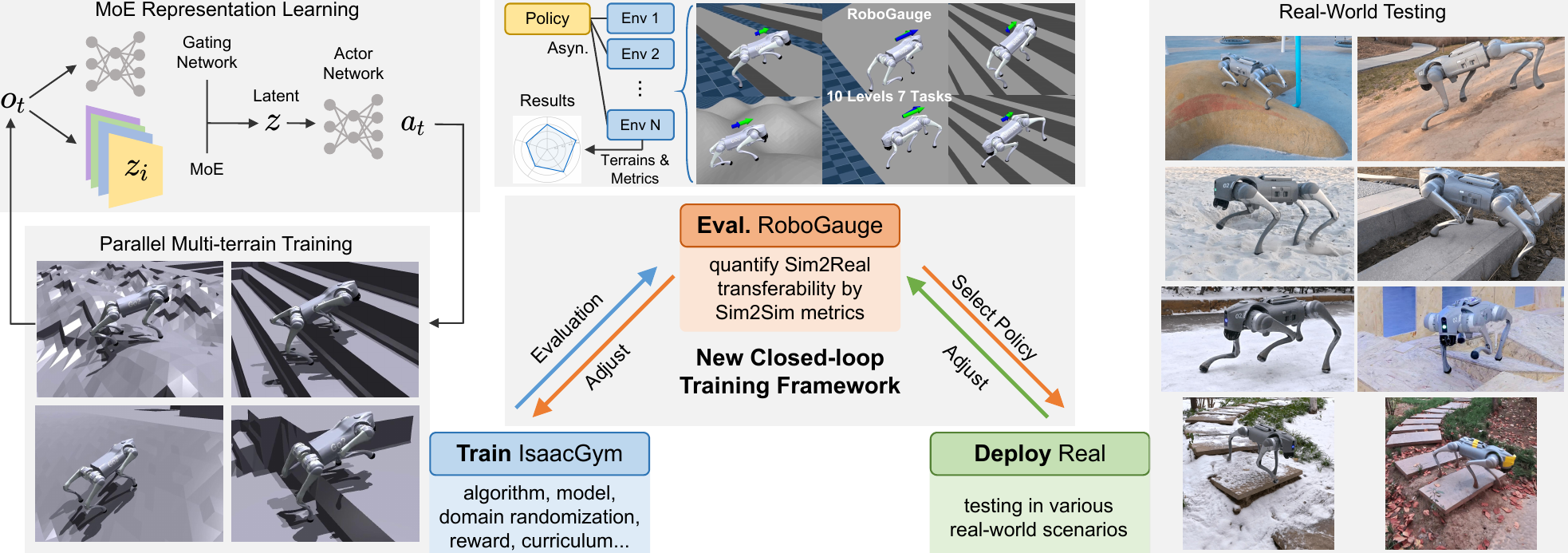}
    \captionof{figure}{
    Our proposed framework integrates a Mixture-of-Experts architecture for terrain and command representation with the RoboGauge assessment suite to quantify sim-to-real transferability through sim-to-sim metrics. This closed-loop design enables reliable policy selection to facilitate robust deployment for agile locomotion across diverse challenging environments based solely on proprioception.
    }
    \label{fig:framework}
    \setcounter{figure}{1}
    \vspace{-0.2cm}
}
\maketitle

\begin{abstract}
Reinforcement learning has shown strong promise for quadrupedal agile locomotion, even with proprioception-only sensing. In practice, however, sim-to-real gap and reward overfitting in complex terrains can produce policies that fail to transfer, while physical validation remains risky and inefficient. To address these challenges, we introduce a unified framework encompassing a Mixture-of-Experts (MoE) locomotion policy for robust multi-terrain representation with RoboGauge, a predictive assessment suite that quantifies sim-to-real transferability. The MoE policy employs a gated set of specialist experts to decompose latent terrain and command modeling, achieving superior deployment robustness and generalization via proprioception alone. RoboGauge further provides multi-dimensional proprioception-based metrics via sim-to-sim tests over terrains, difficulty levels, and domain randomizations, enabling reliable MoE policy selection without extensive physical trials. Experiments on a Unitree Go2 demonstrate robust locomotion on unseen challenging terrains, including snow, sand, stairs, slopes, and 30 cm obstacles. In dedicated high-speed tests, the robot reaches 4 m/s and exhibits an emergent narrow-width gait associated with improved stability at high velocity.
\end{abstract}

\IEEEpeerreviewmaketitle

\begin{figure}[htbp]
    \centering
    \includegraphics[width=0.8\linewidth]{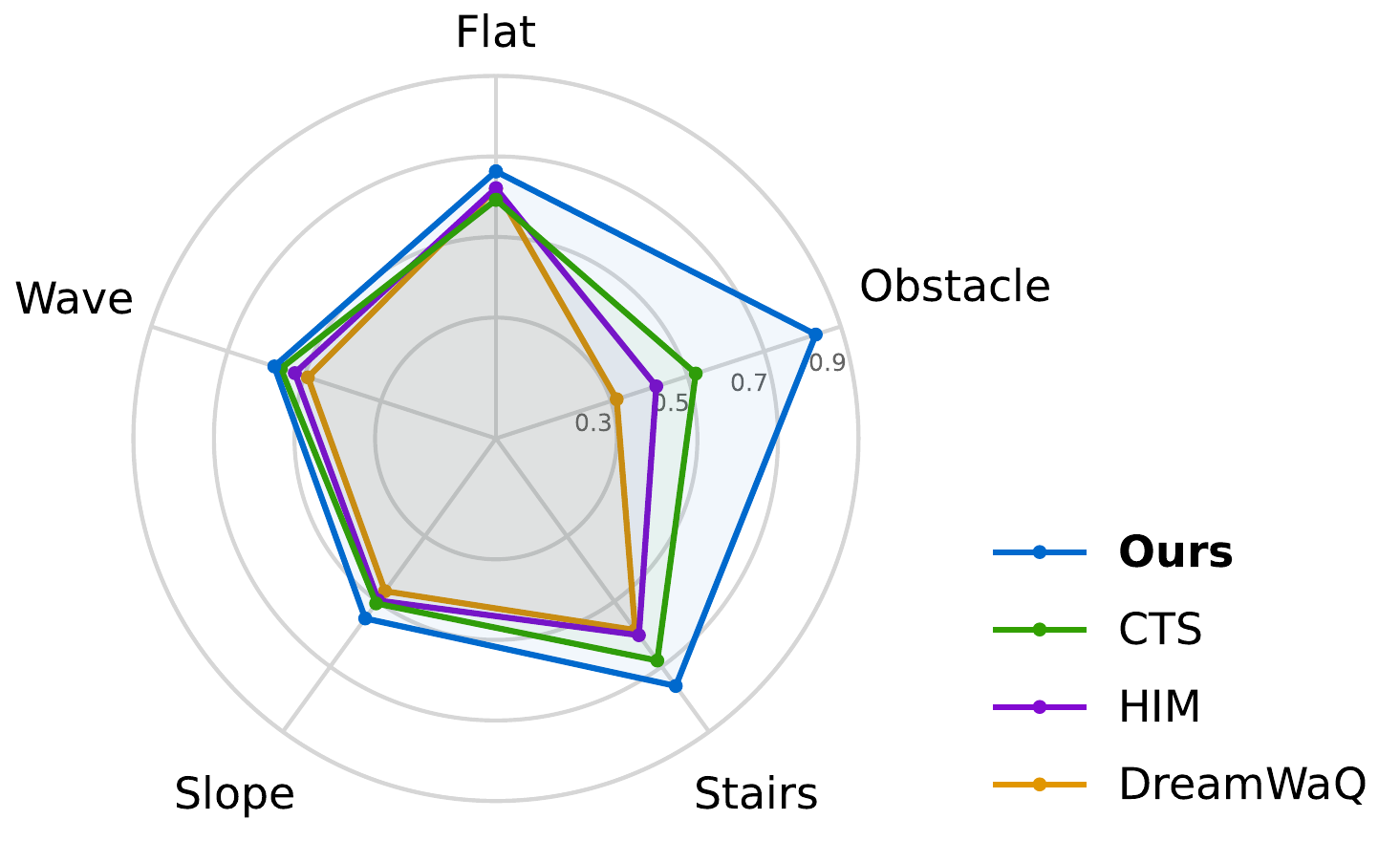}
    \caption{
    Comparative analysis against one-stage proprioceptive methods including CTS, HIM, and DreamWaQ.
    Within the RoboGauge framework, each axis reflects average performance on a specific terrain
    and serves as a reliable proxy to quantify sim-to-real capability.
    Our architecture consistently outperforms or matches previous state-of-the-art across all evaluated terrains
    under RoboGauge's metrics.
    }
    \label{fig:radar}
\end{figure}

\section{Introduction}
Robots frequently operate in complex and dynamic environments which require high levels of mobility \cite{real_time_1986,sampling_based_2011,learning_agile_2019}.
Quadrupedal robots have garnered significant prominence due to their superior mobility and environmental adaptability
\cite{perceptive_whole_body_planning,minimal_human_effort,fine_tuning_real_world,navigation_small_scale_2020robust,collision_free_mpc_2021,collision_free_mpc_2022,state_representation_navigation_2021,vision_aided_dynamic_exploration_2020,walking_in_narrow_spaces_2023,efficient_locally_reactive_controller_2022,vision_guided_quadrupedal_2021,resilient_legged_local_navigation_2024}.
Reinforcement learning has emerged as a potent methodology for motion control by facilitating continuous policy optimization through simulation-based interactions to enhance the robustness of robotic locomotion \cite{eth2020locomotion,kumar2021rma,leveraging_symmetry_rl_locomotion_2024,concurrent_training_2022,learning_robust_agile_locomotion_2023,fine_tuning_real_world,wu2023daydreamer,rudin2022learning,miki2022learning,nahrendra2023dreamwaq,long2024hybrid,margolis2022rapid,fu2022minimizing,margolis2023walk,multi-expert}.


The inherent sim-to-real gap remains a primary barrier as simulation-based performance metrics often prove unreliable for real-world deployment \cite{transferability_2012,domain_randomization_2017,sim_to_real_transfer_2018,dexterous_in_hand_manipulation2020,closing_sim_to_real_loop_2019}.
Specifically, high training rewards across diverse terrains often fail to guarantee physical stability, as policies tend to overfit to the specific dynamics of the simulated robot, thereby degrading generalization to real-world hardware \cite{eth2020locomotion,kumar2021rma,not_only_rewards_2024}.
Moreover, the lack of reliable quantitative proxies compels researchers to rely on direct physical validation, a process that remains prohibitively risky and inefficient \cite{minimal_human_effort,fine_tuning_real_world,wu2023daydreamer}.

To mitigate these challenges,
we propose a training framework that integrates a Mixture-of-Experts (MoE) architecture for terrain and command representation with the RoboGauge assessment suite.
This MoE approach improves modeling capabilities by relying exclusively on proprioception to encode unknown terrains and commands while avoiding exteroceptive sensors like cameras, LiDAR, or foot contact sensors, which frequently fail in extreme conditions such as dense smoke and insufficient lighting or violent shaking.
Complementing the policy iteration architecture we develop RoboGauge as a predictive evaluation framework designed to quantify sim-to-real stability by utilizing a parallelized sim-to-sim methodology across 6 distinct metrics involving 7 terrains and 10 difficulty levels as well as 3 objectives and 4 domain randomizations.

Fig. \ref{fig:radar} illustrates the performance distribution of various models across seven terrains
evaluated within the RoboGauge.
Our MoE policy outperforms all baseline methods across every terrain category to
demonstrate comprehensive superiority.
This approach further exhibits exceptional performance during actual deployment on physical robots.

Our contributions are summarized as follows:
\begin{itemize}
    \item We propose RoboGauge, a comprehensive predictive assessment framework that utilizes a sim-to-sim methodology to quantify sim-to-real transferability, thereby mitigating the risk of hardware damage during direct physical deployment.
    \item We integrate a Mixture-of-Experts module into the policy to resolve existing deficiencies in multi-terrain representation and demonstrate superior mobility on the physical Unitree Go2 robot.
    \item We demonstrate that our framework enables the robot to reach a high-speed locomotion of 4 m/s on flat terrain while exhibiting an emergent narrow-width gait associated with improved stability.
\end{itemize}
\section{Related Work}
\subsection{Reinforcement Learning for Quadrupedal Locomotion}
Reinforcement learning for quadrupedal locomotion in physical environments is hindered by severe sample inefficiency and potential hardware hazards \cite{minimal_human_effort,fine_tuning_real_world,wu2023daydreamer}. The predominant sim-to-real approach employs frameworks such as proximal policy optimization \cite{ppo} or teacher-student training to achieve multi-terrain traversal at velocities under 1 m/s \cite{rudin2022learning, eth2020locomotion, learning-agile-locomotion}. Adaptability has further advanced through latent parameter estimation via adaptation modules or recurrent belief encoders and contrastive learning within parallelized simulations \cite{kumar2021rma, lstm, miki2022learning, nahrendra2023dreamwaq, long2024hybrid}.
Furthermore research pushes agility to peak velocities of 3.9 m/s through command curricula \cite{margolis2022rapid,hacl2025}
whereas diverse gaits \cite{margolis2023walk,gaitor2024,allgaits2025}
and seamless switching emerge from energy optimization rewards \cite{fu2022minimizing,viability2024,non_conflicting2025}
and multi-expert gating architectures \cite{multi-expert,moe-loco}.

\subsection{Sim-to-Real Evaluation Suites}
Evaluation frameworks for locomotion models are currently limited.
In contrast, research in robotic manipulation has addressed similar challenges by employing ranking metrics to verify consistency between simulation and reality \cite{simpler2024,scalable_policy_evaluation_2025}.
High-fidelity digital twins provide closed-loop assessment through environmental reconstruction but often suffer from high costs that restrict their scalability across diverse real-world scenarios \cite{robogsim2024, vrrobo2025}.

\section{MoE Latent Representation Learning}
The proposed one-stage reinforcement learning framework centers on Mixture-of-Experts latent representation learning for quadrupedal locomotion, as illustrated in the training phase of Fig. \ref{fig:framework}.
This section describes the mathematical formulation of the motion control task and the internal structural design of the multi-expert neural network architecture, followed by the detailed reward configurations and environment configurations.

\subsection{Locomotion Control in Reinforcement Learning}
The core objective of quadrupedal locomotion control is to determine appropriate joint torque commands
for all actuated joints based on proprioception.
Assuming that proprioceptive information is acquired exclusively via an IMU and joint encoders,
the quadrupedal locomotion dynamics are modeled as an infinite-horizon Partially Observable Markov Decision Process (POMDP),
defined by the tuple $(\mathcal{S}, \mathcal{A}, \mathcal{O}, P, \Omega, R, \rho_0)$,
where $\mathcal{S}\subset\mathbb{R}^n$ denotes the privileged state space including all dynamic information
of robot perception and the surrounding environment.
The set $\mathcal{A}\subset\mathbb{R}^m$ represents the action space and $\mathcal{O}\subset\mathbb{R}^o$ signifies the observation space.
The state transition probability is characterized by $P(\boldsymbol{s}'|\boldsymbol{s},\boldsymbol{a})$,
the observation function by $\Omega(\boldsymbol{o}|\boldsymbol{s})$,
the reward function by $R(\boldsymbol{s},\boldsymbol{a},\boldsymbol{s}')$,
and the initial state distribution by $\rho_0(\boldsymbol{s}_0)$.
Our objective is to acquire an optimal policy $\pi^*$ that maximizes the expected cumulative discounted reward over the trajectory $\tau=\{\boldsymbol{s}_t,\boldsymbol{a}_t,r_t,\boldsymbol{s}_{t+1},...\}$:
\begin{equation}
    J(\pi)=\mathbb{E}_{\boldsymbol{s}_0\sim\rho_0,\tau\sim\pi}\left[\sum_{t=0}^{\infty}\gamma^t R(\boldsymbol{s}_t,\boldsymbol{a}_t,\boldsymbol{s}_{t+1})\right]
\end{equation}
where $\gamma\in(0,1)$ serves as the discount factor.

Let $\boldsymbol{o}_t \in \mathcal{O}$ and $\boldsymbol{s}_t \in \mathcal{S}$ denote the observation and state at time $t$, respectively.
The observation incorporates the angular velocity $\boldsymbol{\omega}$ measured by the IMU,
the projected gravity vector $\boldsymbol{g}_{\text{proj}}$ in the body frame,
joint positions $\boldsymbol{q}$, and joint velocities $\boldsymbol{\dot{q}}$,
linear velocity commands in the longitudinal and lateral directions $v_x^{\text{cmd}}$ and $v_y^{\text{cmd}}$,
the yaw rate command $\omega_z^{\text{cmd}}$, and the preceding action $\boldsymbol{a}_{t-1}$.
Beyond the components of $\boldsymbol{o}_t$, the state $\boldsymbol{s}_t$ encompasses the linear velocity $\boldsymbol{v}_t$,
sampled terrain heights $\boldsymbol{h}_t$, and environmental latent parameters $\boldsymbol{\mu}_t$ representing foot contact forces,
joint torques, and joint accelerations.
The height measurements are sampled within a $1\text{m} \times 1.6\text{m}$ rectangular area
centered on the robot's base with a $0.1\text{m}$ interval, providing a comprehensive representation of the local terrain.

The action $a_t \in \mathcal{A}$ denotes the joint position offsets relative to the initial joint positions.
For each actuated joint, the model produces target positions,
and the required torques are computed through a Proportional-Derivative (PD) controller.

\subsection{Mixture-of-Experts Representation Encoder}
To facilitate the acquisition of an optimal policy,
privileged observations $\boldsymbol{s}_t$ are commonly employed during training to
accelerate learning and elevate performance upper bounds.
Given that the model is restricted to observations $\boldsymbol{o}_t$
during deployment, the teacher-student paradigm leverages distillation techniques to
transfer advantageous strategies to the student \cite{eth2020locomotion}.
The Concurrent Teacher-Student (CTS) framework \cite{wang2024cts} simultaneously optimizes
both teacher and student networks. Through this parallel learning process,
both entities update actor and critic networks,
enabling student feedback to actively refine the teacher's parameters.
Such joint optimization typically yields outcomes superior to those achieved
through independent training \cite{deep_mutual_learning}.
We observe that the limited expressive capacity of the student model
often precludes it from accurately inferring the features encoded by the teacher,
which consequently restricts the performance ceiling.
To overcome this limitation, we integrate a Mixture-of-Experts (MoE) structure \cite{moe1, moe2}
into the student architecture within the CTS framework.
This augmentation bolsters the representational capabilities of
the student and further elevates the performance upper bound of the overall system.

We substitute the student encoder in the CTS framework
with the MoE network.
This architecture comprises $K$ parallel expert subnetworks $\{E_k\}_{k=1}^K$
where each expert specializes in processing observation data under specific command types or environmental contexts.
To coordinate these subnetworks,
we incorporate a gating network $g$ that dynamically allocates weights $\omega_k$
based on the observation sequence $\boldsymbol{o}_{t-H:t}=\left[\boldsymbol{o}_{t-H}, \cdots, \boldsymbol{o}_t\right]^T$.
These coefficients determine the relative contribution of each expert
to the current state representation.
Accordingly, the resulting latent state $\boldsymbol{z}_s$ of the student encoder
is formulated as the weighted sum of all expert outputs:
\begin{equation}
    \boldsymbol{z}_s = \sum_{k=1}^{K}\omega_k E_k(\boldsymbol{o}_{t-H:t}),\quad \omega_k = \text{softmax}(g(\boldsymbol{o}_{t-H:t}))_k
\end{equation}

To prevent the gating network from exclusively activating a single expert subnetwork,
we incorporate an auxiliary load balancing loss \cite{switch_transformer, sparsely_gated_moe}:
\begin{equation}
    \mathcal{L}_{\text{load balance}} = \sum_{k=1}^K\left(\bar{\omega}_k-\frac{1}{K}\right)^2,\quad \bar{\omega}_k = \frac{1}{B}\sum_{j=1}^B\omega_k^{(j)}
\end{equation}
where $B$ specifies the batch size utilized during training
while $\omega_k^{(j)}$ represents the weight allocated to the $k$-th expert for the $j$-th sample.
This formulation encourages the system to distribute tasks uniformly
across all experts to ensure representational diversity and expressive capacity.

\subsection{Reward Design}
We utilize a consistent reward function structure for both the multi-terrain and
the flat-ground high-speed locomotion models.
The fundamental reward configurations are established based on
established methodologies \cite{eth2020locomotion, rudin2022learning, wang2024cts}.
Building upon these foundations,
we introduced a hip joint position reward to mitigate outward thigh abduction during rapid locomotion.
Appendix Table~\ref{table:reward} presents the comprehensive reward specifications.
Within this framework, $\sigma$ denotes the velocity tracking precision parameter initialized to a value of 0.25.
Additionally, the reward component $r^{\text{fr}}$ adopts the formulation from the CTS model \cite{wang2024cts}
to incentivize adequate foot clearance during high-speed movement.
For high-speed locomotion training on flat ground,
we introduce an external hip symmetry reward $r^{\text{hs}}$ to regularize joint positions
while executing longitudinal linear motion commands.
This term ensures that the robot maintains symmetrical postures and is defined as follows:
\begin{equation}
    r^{\text{hs}} = \frac{|v_{x}^{\text{cmd}}|}{\|\boldsymbol{v}^{\text{cmd}}\|_2} \cdot \left( |q_{\text{FL}}^{\text{hip}} + q_{\text{FR}}^{\text{hip}}| + |q_{\text{RL}}^{\text{hip}} + q_{\text{RR}}^{\text{hip}}| \right)
\end{equation}

Since the training curriculum involves diverse terrains,
the vertical linear velocity reward weight decays to zero once the robot achieves stable locomotion.
This reduction prevents vertical velocity fluctuations caused by terrain irregularities from
interfering with the policy optimization process.
We observed that augmenting the base height reward weight effectively mitigates body sagging
during high-speed locomotion on flat surfaces.
For the multi-terrain model, the reference base height is established at 0.38m.
In contrast, the high-speed model utilizes a lower reference height of 0.33m to
enhance the stability of the center of mass through a reduced posture.

\subsection{Environment Configurations}

We utilize the IsaacGym simulation environment \cite{isaacgym} to train 8192 agents in parallel across diverse terrains.
The experimental platform is the Unitree Go2 quadrupedal robot featuring 12 degrees of freedom.
Motor PD control gains are specified as $k_\text{p}=20.0$ and $k_\text{d}=0.5$ for all joints.
The system operates with a control frequency of 50Hz and a simulation frequency of 200Hz.
The length of the observation sequence $\boldsymbol{o}_{t-H:t}$ is set to 5 for MoE input.
Algorithm configurations follow the CTS framework \cite{wang2024cts}.

Establishing a proper curriculum difficulty is essential to ensure representational diversity during training.
Following \cite{rudin2022learning},
we implement a terrain curriculum encompassing seven terrains including
flat, wave, slope, rough slope, stairs up, stairs down, and obstacle.
Slope inclinations vary from $5.7^\circ$ to $29.6^\circ$ and the rough slope terrain incorporates random height fluctuations of 5cm.
Stair heights range between $5\text{cm}$ and $25.7\text{cm}$ with a constant tread width of $31\text{cm}$.
The obstacle terrain consists of random cubic structures with heights spanning from $5\text{cm}$ to $27.5\text{cm}$
and widths between $1\text{m}$ and $2\text{m}$.

To facilitate effective sim-to-real transfer, we introduce domain randomization parameters,
the details of which are shown in Table~\ref{table:domain_randomization}.

\begin{table}[!h]
\centering
\caption{Domain Randomization Specifications} \label{table:domain_randomization}
\begin{tabularx}{\linewidth} {
   >{\arraybackslash}X 
   >{\arraybackslash\hsize=1.2\hsize}X
   >{\arraybackslash\hsize=.5\hsize}X
  }
\toprule
\textbf{Randomization Term} & \textbf{Range} & \textbf{Unit} \\
\midrule
Friction & $[0.5, 1.5]$ & -- \\
Payload mass & $[-1, 1]$ & kg \\
Link mass & $[0.9, 1.1] \times \text{Nominal Value}$ & kg \\
Base center of mass & $[-3, 3] \times [-3, 3] \times [-3, 3]$ & cm \\
Restitution & $[0.0, 0.5]$ & -- \\
Proportional gain $k_{\text{p}}$ & $[0.9, 1.1] \times \text{Nominal Value}$ & $\text{Nm}/\text{rad}$ \\
Derivative gain $k_{\text{d}}$ & $[0.9, 1.1] \times \text{Nominal Value}$ & $\text{Nm}\cdot\text{s}/\text{rad}$ \\
Actuator strength & $[0.8, 1.2] \times \text{Nominal Value}$ & -- \\
Actuator offset & $[-0.035, 0.035]$ & rad \\
Control latency & $[0, 20]$ & ms \\
\bottomrule
\end{tabularx}
\end{table}

We identify several training problems within the original framework \cite{rudin2022learning,nahrendra2023dreamwaq,long2024hybrid,wang2024cts} which are elaborated in Appendix~\ref{appendix:training_details} along with corresponding ablation studies to verify the effectiveness of our improvements. To ensure reward stability on complex terrains we implement a \textit{dynamic velocity tracking precision adjustment} \ref{appendix:dynamic_sigma} that scales constraints based on terrain difficulty and command magnitude.
We further incorporate a comprehensive command design suite including a \textit{command curriculum}, \textit{extreme command sampling} and \textit{dynamic command sampling} \ref{appendix:command_design} to ensure consistent progression through terrain levels.
These strategies collectively accelerate convergence and elevate the peak RoboGauge score by 11\% while promoting stable locomotion patterns across diverse environments.

\section{The RoboGauge Predictive Assessment Framework}
As illustrated in the central evaluation module of Fig.~\ref{fig:framework}, RoboGauge serves as the pivotal assessment engine designed to bridge the gap between simulation training and real-world deployment.
This section details the design philosophy of RoboGauge, a comprehensive framework developed to quantitatively validate the performance of reinforcement learning (RL) locomotion controllers.

\begin{figure}[htbp]
    \centering
    \includegraphics[width=0.95\linewidth]{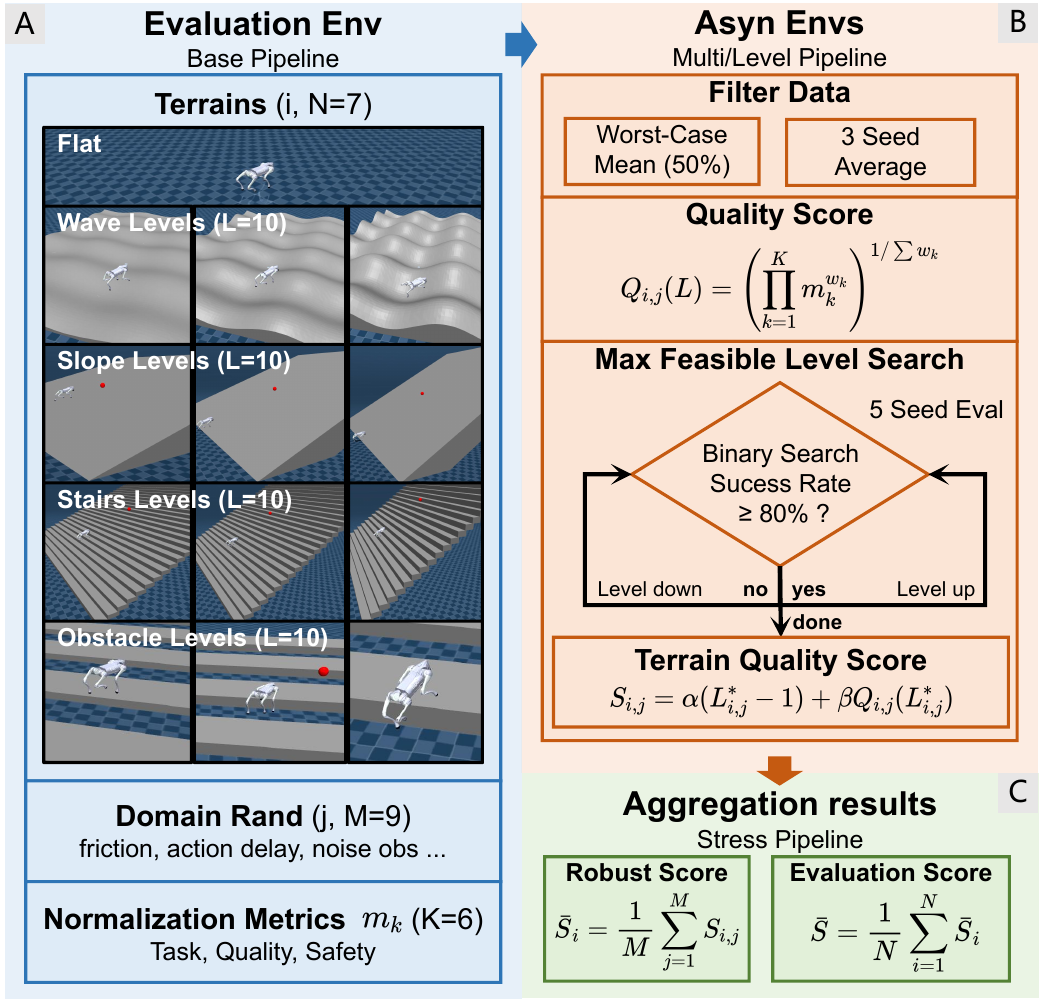}
    \caption{The RoboGauge evaluation architecture consists of three hierarchical stages.
    (\textbf{A}) Base Pipeline serves as a single evaluation environment
    by incorporating specific terrains and domain randomization.
    (\textbf{B}) Multi/Level Pipeline highlights the parallel evaluations across diverse random seeds.
    (\textbf{C}) Stress Pipeline triggers comprehensive testing across the entire terrain suite to synthesize the final score.}
    \label{fig:robogauge}
\end{figure}

Built upon the MuJoCo~\cite{mujoco} simulation environment, the framework's operational workflow is depicted in Fig.~\ref{fig:robogauge}, which organizes the evaluation process into three hierarchical stages:
(1) the BasePipeline for atomic, single-environment evaluations;
(2) the Multi/Level Pipeline for parallelized difficulty assessment and domain randomization;
and (3) the Stress Pipeline for synthesizing a unified robustness score.
The following subsections detail the formulation of our quantitative metrics, the design of the evaluation environments, and the hierarchical scoring methodology, respectively.

\begin{table}[!h]
\centering
\caption{Metrics for the RoboGauge Framework}
\label{table:metrics} 
\begin{tabularx}{\linewidth}{
    >{\raggedright\arraybackslash\hsize=.7\hsize}X 
    >{\raggedright\arraybackslash\hsize=1.3\hsize}X 
}
\toprule
\textbf{Metric} & \textbf{Description} \\ \midrule
Lin. Velocity Error & Linear velocity $\ell_2$ tracking error \\
Ang. Velocity Error & Angular velocity $\ell_2$ tracking error \\
Dof Power & Motor power consumption \\
Dof Limits & Joint angles exceeding soft limits \\
Orientation Stability & Gravity projection on the lateral ($y$) axis \\
Torque Smoothness & Temporal smoothness of motor torques \\
ZMP Margin & Normalized Zero Moment Point deviation \\
Friction Margin & Normal-force-weighted Coulomb friction \\ \bottomrule
\end{tabularx}
\end{table}

\subsection{Quantitative Performance Metrics}
The primary objective of RoboGauge is to derive quantitative indicators solely from proprioceptive feedback that accurately reflect a controller's efficacy during real-world deployment.
Drawing from empirical observations of common failure modes in physical testing, we formulated 6 metrics, as detailed in Table~\ref{table:metrics}, addressing three critical aspects of sim-to-real transfer.
First, to ensure \textbf{hardware safety and efficiency}, we evaluate \textit{dof limits} and \textit{dof power}, preventing actuator damage or thermal failure caused by sub-optimal motor operation.
Second, \textbf{tracking precision} is quantified by the \textit{velocity error}, measuring the controller's fidelity in following linear and angular commands.
Finally, we assess \textbf{motion stability} via \textit{torque smoothness} and \textit{orientation stability} to mitigate structural vibrations and ensure robust attitude control. To further formalize this stability assessment, we integrated two physical criteria: the Zero Moment Point (ZMP) margin \cite{ZMP2004} and a Coulomb friction margin under Contact Wrench Cone (CWC) constraints \cite{CWC2015}. The ZMP margin evaluates the horizontal distance error of the ZMP relative to the nominal stance span, derived via aggregated Newton-Euler equations. The Coulomb friction margin computes the normal-force-weighted average slack to the friction-cone boundary over active contacts. Detailed mathematical formulations for these stability metrics are provided in Appendix \ref{appendix:stability_metric}.
To facilitate a unified assessment, all raw measurements are normalized and transformed via the function $f(x)=1-x$, ensuring that a higher score consistently signifies superior performance.

\subsection{Evaluation Environment and Randomization}
To ensure a rigorous and holistic assessment, the framework establishes a systematic evaluation matrix integrating diverse motion goals, complex terrain structures, and extensive domain randomizations.

\subsubsection*{Motion Goals}
We devised motion goals to stress-test the control policy as detailed in Appendix Table~\ref{table:goals}.
These tasks cover maximum command execution, rapid emergency stops, and abrupt diagonal velocity step changes.
Furthermore, the evaluation incorporates a specific target position task regulated by a proportional error controller.
This task serves as the pass criterion for terrain traversal.
It enables a binary search strategy to identify the maximum difficulty level the model can navigate.

\subsubsection*{Terrain Configuration}
The evaluation suite features 5 distinct terrain categories: flat, wave, slopes, stairs, and obstacles.
Excluding the flat surface, each terrain type is subdivided into 10 discrete difficulty levels to probe the limits of the controller's mobility.
Fig.~\ref{fig:robogauge} explicitly illustrates the environmental complexity for difficulty levels 3, 5, and 10.
Beyond difficulty scaling, navigation on slopes and stairs presents unique directional challenges.
Therefore, we explicitly evaluate both ascending and descending configurations to ensure robust performance regardless of the incline direction.

\subsubsection*{Domain Randomization}
We implement domain randomization across two primary dimensions environmental factors and inherent robot properties.
Specifically, environmental factors include variations such as payloads and friction coefficients, while robot properties encompass motor response latency and observation noise.
Collectively, these perturbations simulate the imperfections of physical hardware, preventing the policy from overfitting to ideal simulation dynamics and ensuring robust real-world transfer.

\subsection{Hierarchical Scoring Methodology}

We denote the set of $N=7$ terrain configurations as $\mathcal{T}=\{T_1,\dots,T_N\}$, expanding the five distinct terrain categories by treating ascending and descending directions on slopes and stairs as separate evaluation environments.
For each terrain $T \in \mathcal{T}$, we apply $M=9$ distinct domain randomizations, denoted by $\mathcal{D}=\{d_1,\dots,d_M\}$.
The terrain difficulty is stratified into $10$ levels, represented as $L\in\{1,2,\dots,10\}$.
Each evaluation session yields $K=8$ performance metrics, designated as $\mathcal{M}=\{m_1,\dots,m_K\}$.

Next, we formalize the composite scoring methodology for evaluating the model.
For a given terrain $T_i$, domain randomization $d_j$, and difficulty level $L$,
we aggregate $K=8$ normalized metrics $\{m_1, \dots, m_8\}$,
where each $m_k \in [0,1]$ denotes the average result across three stochastic seeds.
To penalize imbalanced performance, specifically to prevent high scores
when a critical dimension fails, we employ a weighted geometric mean to compute the execution quality score:
\begin{equation}\label{eq:quality_score}
Q_{i,j}(L)=\left(\prod_{k=1}^Km_k^{w_k}\right)^{1/\sum_{k=1}^Kw_k}
\end{equation}

We adopt a Worst-Case Mean aggregation strategy to evaluate performance across motion goals.
This method involves averaging the lowest 50\% of scores within each goal, effectively discounting high scores from non-challenging commands to concentrate the assessment on challenging maneuvers such as obstacle negotiation and gait transitions.
Additionally, we compute the global mean and the average of the top 25\% for broader reference as detailed in Appendix Table \ref{table:detailed-baseline-metrics}.

We employ a binary search strategy to identify the maximum attainable difficulty level $L^*_{i,j} \in \mathcal{L}$ for each terrain under the specified domain randomization parameters.
For a given level, the model is evaluated across five stochastic seeds to verify whether it successfully reaches the goal.
A difficulty level is deemed passable if the success rate in the goal-reaching task surpasses 80\%.

Let $Q_{i,j}(L^*_{i,j})$ denote the execution quality score at the highest passable difficulty level.
To balance task difficulty and execution quality across diverse terrains,
the terrain quality score $S_{i,j}$ for a specific terrain $T_i$ and
domain randomization $d_j$ is formulated using the following overlapping scoring function:
\begin{equation}\label{eq:overlapping_scoring_function}
S_{i,j}=\alpha (L_{i,j}^*-1)+\beta Q_{i,j}(L_{i,j}^*)
\end{equation}
By setting $\beta > \alpha$, this design ensures that high-quality performance at a lower difficulty level approximates the score of mediocre performance at a higher level, facilitating a smooth transition across difficulty tiers.

The framework results are aggregated through arithmetic averaging.
Initially, we calculate the robust score $\bar{S}_i$ for each terrain $T_i$ by averaging the results over $M$ domain randomizations.
The final framework score $\bar{S}$ is subsequently obtained by averaging these robust scores across all $N$ terrains:
\begin{equation}
\bar{S}_{i}=\frac{1}{M}\sum_{j=1}^{M}S_{i,j}, \quad \bar{S}=\frac{1}{N}\sum_{i=1}^{N}\bar{S}_{i}
\end{equation}


Given the extensive combinations of terrain types, randomization parameters, and random seeds, performing a full evaluation sequentially is prohibitively time-consuming.
We consequently adopt multiprocessing acceleration to run concurrent environment instances.
This efficiency fulfills the necessity for rapid performance feedback throughout the training phase.
Further implementation specifics and all specific hyperparameter values are elaborated in Appendix~\ref{appendix:robogauge}.

\section{Framework Validation and Ablation Studies}
In this section, we present experiments aimed at addressing the following research questions:
\begin{itemize}
    \item \textbf{Q1}: Does RoboGauge provide metrics that correlate closely with real-world performance?
    \item \textbf{Q2}: How do state-of-the-art methods perform under our evaluation framework?
    \item \textbf{Q3}: Can the Mixture of Experts architecture effectively differentiate between various encoded terrains?
\end{itemize}
\subsection{Metric Reliability of RoboGauge}
We deployed the proposed model and baselines on a Unitree Go2 quadruped robot. We utilize a 12-camera NOKOV Mars18H motion capture system operating at 90Hz to acquire real-time linear and angular velocity data across flat terrain and 10cm stairs by mounting five markers on the robot base. 
At the same time, we gather proprioceptive feedback and motor torques to derive the six specific metrics in Table~\ref{table:metrics}. 
To quantify the fidelity of these assessment methods, we compare the metric errors from both the training environment and our proposed framework against real-world ground truth. 
We specifically evaluate a model that exhibited high performance during training but suffered from significant sim-to-real degradation. 
As presented in Table \ref{table:metrics_error}, the training environment consistently yields larger errors. 
Comprehensive scoring data provided in Table \ref{table:full_comparison_data} in the Appendix further confirms that errors obtained through our framework are markedly lower than those from standard training evaluations. 
These results demonstrate that our evaluation framework more accurately reflects real-world performance and provides a more dependable basis for model selection.

\begin{table}[!h]
\centering
\caption{Metrics Error Comparison}
\label{table:metrics_error}
\begin{tabularx}{\linewidth} {
   >{\arraybackslash\hsize=.8\hsize}X 
   >{\centering\arraybackslash\hsize=1.0\hsize}X
   >{\centering\arraybackslash\hsize=1.2\hsize}X
   >{\centering\arraybackslash\hsize=.9\hsize}X
   >{\centering\arraybackslash\hsize=1.0\hsize}X
  }
\toprule
\textbf{Env.} & \textbf{Cmd.} & \textbf{Tracking} $\downarrow$ & \textbf{Safety} $\downarrow$ & \textbf{Quality} $\downarrow$ \\
\midrule
\multirow{4}{*}{\shortstack{MuJoCo \\ (\textbf{Ours})}} & Longitudinal & 0.0573 & 0.0253 & 0.0246 \\
                          & Lateral      & 0.0541 & 0.0049 & 0.0079 \\
                          & Angular      & 0.0560 & 0.0050 & 0.0035 \\
                          \cmidrule(lr){2-5}
                          & \textbf{Average} & \textbf{0.0558} & \textbf{0.0117} & \textbf{0.0120} \\
\midrule
\multirow{4}{*}{\shortstack{IsaacGym \\ (Training)}} & Longitudinal & 0.1365 & 0.0844 & 0.0678 \\
                                                        & Lateral      & 0.0572 & 0.0052 & 0.0125 \\
                                                        & Angular      & 0.0713 & 0.0103 & 0.0337 \\
                                                        \cmidrule(lr){2-5}
                                                        & \textbf{Average} & 0.0883 & 0.0333 & 0.0380 \\
\bottomrule
\end{tabularx}
\end{table}

\subsection{Comparison of Baselines under RoboGauge}
To facilitate a rigorous comparative evaluation, we benchmark our proposed approach against several state-of-the-art one-stage training algorithms based solely on proprioception:
\begin{enumerate}
    \item DreamWaQ \cite{nahrendra2023dreamwaq}: The policy utilizes an asymmetric actor-critic scheme with a variational estimator to jointly predict body velocity and terrain latents.
    \item HIM \cite{long2024hybrid}: The policy incorporates a hybrid internal model to explicitly estimate robot responses using contrastive learning.
    \item CTS \cite{wang2024cts}: The policy employs an asymmetric teacher-student setup to optimize the agent via reinforcement learning and supervised reconstruction.
\end{enumerate}

We implement all aforementioned methods using a consistent configuration, with 8192 parallel agents training in IsaacGym \cite{isaacgym}.
Because DreamWaQ and HIM do not support terrain-specific velocity command ranges, we set their maximum limit to 1 m/s. We apply this same constraint within the RoboGauge assessment for these models to reduce the difficulty of command tracking. Conversely, both CTS and our proposed model utilize a command range of 2 m/s for both training and evaluation. Each algorithm is trained with three independent random seeds and we select the model achieving the highest RoboGauge score for subsequent analysis. The outcomes summarized in Table~\ref{table:baseline} demonstrate that our method significantly outperforms the other approaches across the entire set of metrics.
\begin{table}[!h]
\centering
\caption{RoboGauge results for baselines} \label{table:baseline}
\begin{tabularx}{\linewidth} {
   >{\arraybackslash\hsize=1.2\hsize}X 
   >{\centering\arraybackslash\hsize=.5\hsize}X
   >{\centering\arraybackslash\hsize=1.2\hsize}X
   >{\centering\arraybackslash\hsize=1.0\hsize}X
   >{\centering\arraybackslash\hsize=1.0\hsize}X
   >{\centering\arraybackslash\hsize=0.3\hsize}X
  }
\toprule
\textbf{Model} & \textbf{Score} & \textbf{Tracking} $\uparrow$ & \textbf{Safety} $\uparrow$ & \textbf{Quality} $\uparrow$ & \textbf{Level} \\
\midrule
\textbf{Ours} & \textbf{0.6713} & \textbf{0.6669} & \textbf{0.7857} & \textbf{0.7392} & \textbf{7.85} \\
CTS & 0.5786 & 0.5755 & 0.7066 & 0.6624 & 6.83 \\
HIM & 0.5379 & 0.5453 & 0.6476 & 0.6050 & 6.19 \\
DreamWaQ & 0.5054 & 0.5105 & 0.6149 & 0.5730 & 5.74 \\
\bottomrule
\end{tabularx}
\end{table}

\begin{figure}[htbp]
    \centering
    \begin{minipage}{\linewidth}
        \includegraphics[width=\linewidth]{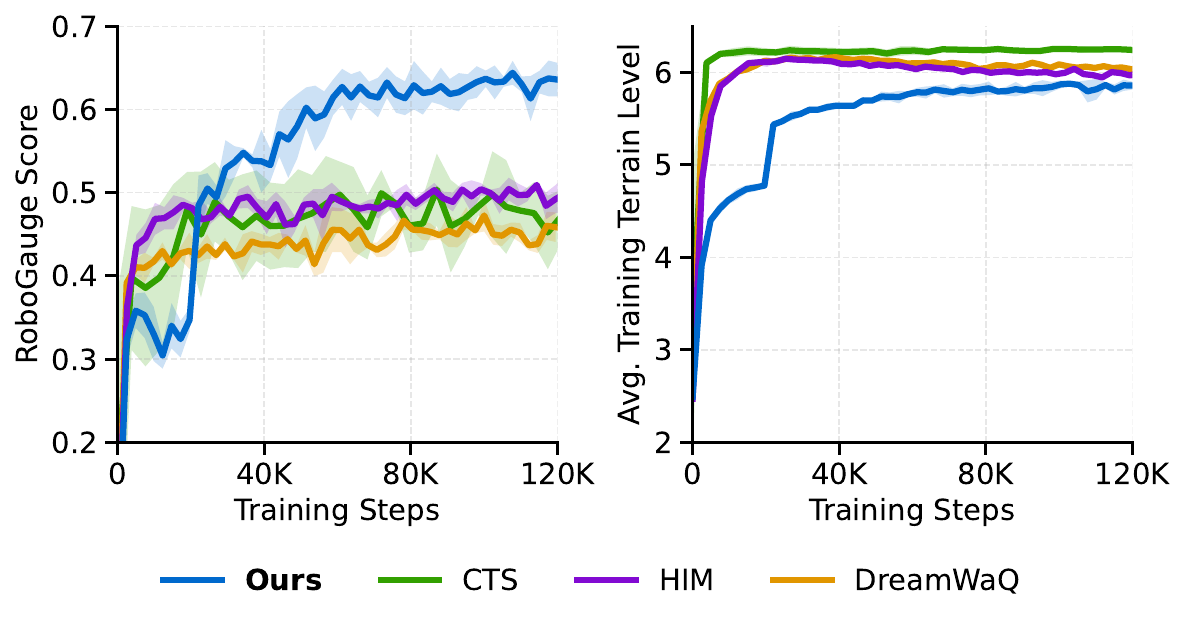}
        \caption{Comparison of RoboGauge scores and terrain level curves across various baselines during training. 
        Stable RoboGauge scores despite fluctuating terrain levels demonstrate that training levels fails to accurately represent model performance.}
        \label{fig:training_curves_baseline}
    \end{minipage}

    \vspace{0.3cm}

    \begin{minipage}{\linewidth}
        \centering
        \includegraphics[width=\linewidth]{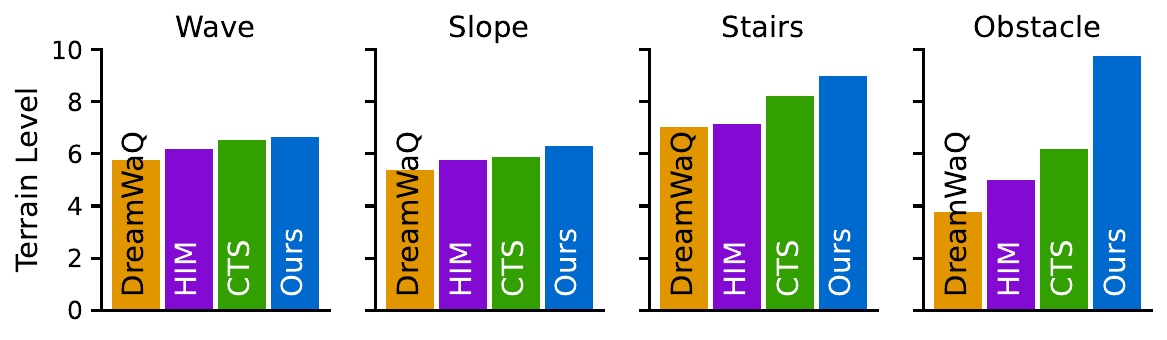}
        \caption{Comparison of maximum terrain levels across varying friction coefficients as evaluated by RoboGauge.}
        \label{fig:level_vs_friction}
    \end{minipage}
    \vspace{-0.2cm}
\end{figure}

As indicated in the training curves in Fig. \ref{fig:training_curves_baseline}, our model does not necessarily achieve the highest terrain levels during the training phase compared to other baselines. 
Nevertheless, the predictability assessment framework provides precise scores that accurately reflect the underlying performance. 
Fig. \ref{fig:level_vs_friction} illustrates the maximum terrain levels attained across a variety of friction coefficients. Details of the terrain levels are provided in Fig. \ref{fig:level_vs_friction_full} of the Appendix.
Our model consistently exhibits superior terrain level proficiency across the entire range of friction values. 
These findings are further corroborated by the real-world deployment data in Table \ref{table:survival_rate_optimized}, which confirms that the controller possesses the capability to navigate such challenging environments in physical settings.

\subsection{Ablation and Latent Representation of MoE}

We designed various ablation studies to investigate the integrated MoE structure, including the following variants:
\begin{enumerate}
    \item MoE-NG: The command information is excluded from the MoE input, utilizing only observation information to the expert networks.
    \item AC-MoE: Following MoE-Loco \cite{moe-loco}, the MoE structure is applied to the Actor-Critic networks rather than the student encoder.
    \item MCP \cite{mcp}: A multiplicative composition strategy is employed for the actions output by the Actor.
\end{enumerate}

As shown in Table \ref{table:ablation}, our proposed method achieved the best performance across all evaluation metrics. Furthermore, during training, we observed that modifications to the action network, such as AC-MoE and MCP, were prone to loss divergence.
This instability likely originates from the expert combination acting directly within the action space. The concurrent adaptation of the gating network and individual experts can yield volatile control signals that induce hazardous maneuvers and consequently undermine training stability.

\begin{table}[!h]
\centering
\caption{RoboGauge Results for MoE Ablation} \label{table:ablation}
\begin{tabularx}{\linewidth} {
   >{\arraybackslash\hsize=1.4\hsize}X
   >{\centering\arraybackslash\hsize=.8\hsize}X
   >{\centering\arraybackslash\hsize=.8\hsize}X
   >{\centering\arraybackslash\hsize=.8\hsize}X
   >{\centering\arraybackslash\hsize=.8\hsize}X
   >{\centering\arraybackslash\hsize=0.6\hsize}X
  }
\toprule
\textbf{Model} & \textbf{Score}    & \textbf{Tracking} & \textbf{Safety}   & \textbf{Quality}  & \textbf{Level}  \\
\midrule
\textbf{MoE (Ours)} & \textbf{0.6713} & \textbf{0.6669} & \textbf{0.7857} & \textbf{0.7392} & \textbf{7.85} \\
AC-MoE \cite{moe-loco} & 0.6509 & 0.6442 & 0.7644 & 0.7149 & 7.52 \\
MoE-NG & 0.6519 & 0.6447 & 0.7639 & 0.7186 & 7.56 \\
MCP \cite{mcp} & 0.6399 & 0.6355 & 0.7542 & 0.7058 & 7.41 \\
\bottomrule
\end{tabularx}
\end{table}

\begin{figure}[htbp]
    \centering
    \begin{minipage}{\linewidth}
        \includegraphics[width=\linewidth]{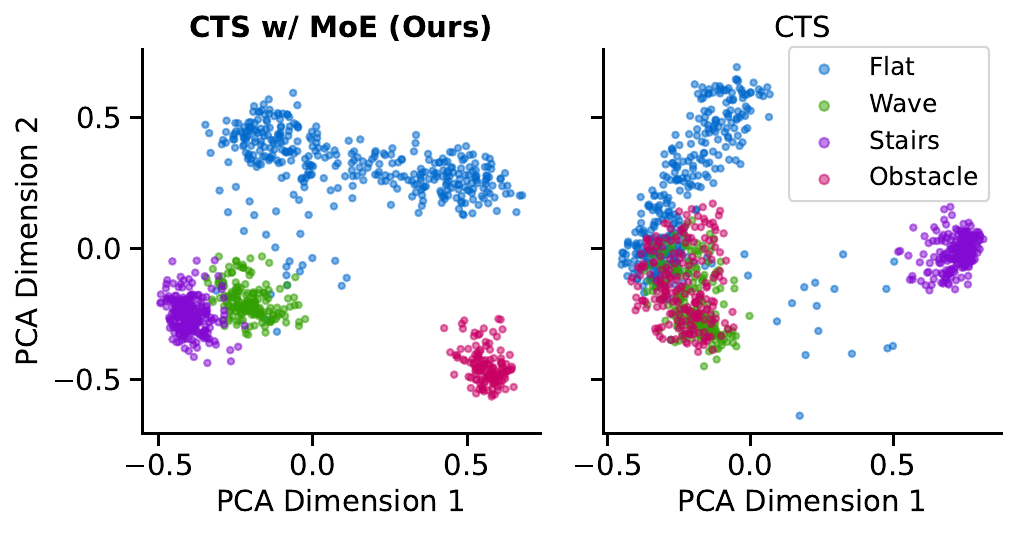}
        \caption{PCA visualization of the student encoder latent space in different terrains with forward command.}
        \label{fig:pca_plot}
    \end{minipage}
\end{figure}

We subsequently visualize the MoE latent space by applying Principal Component Analysis \cite{pca} to reduce the dimensionality of the student encoder hidden states.
Fig. \ref{fig:pca_plot} contrasts the state distributions during 5~s of forward locomotion across diverse terrains to evaluate the impact of the MoE module. Similarly, Fig. \ref{fig:cmd_pca_plot} in the Appendix illustrates the hidden state distributions across all terrains under various commands including forward, backward, left, and right turns over a 5~s duration. 
These results indicate that the MoE architecture achieves superior discrimination of encoding features across various terrains and motion commands.

\section{Physical Deployment and Generalization}
In this section, our real-world experiments are designed to address the following research questions.
\begin{itemize}
    \item \textbf{Q4}: Does the proposed framework outperform more challenging terrain compared to other baselines?
    \item \textbf{Q5}: How accurate is its tracking of velocity commands?
    \item \textbf{Q6}: Can the model perform reliably in diverse complex environments not encountered during training? 
\end{itemize}

\begin{figure*}
    \centering
    \begin{tikzpicture}
        \node[anchor=south west, inner sep=0] (image) at (0,0) {\includegraphics[width=\textwidth]{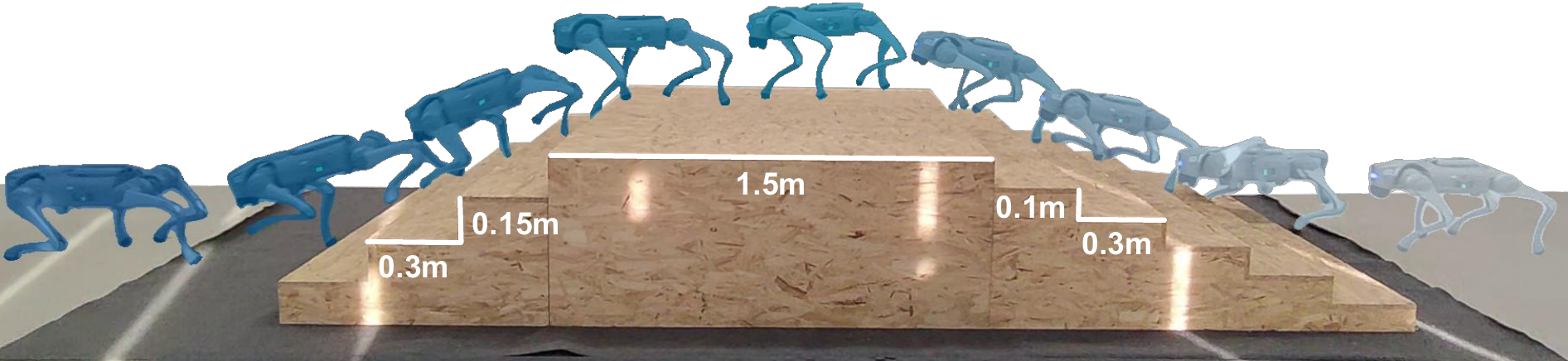}};
        \begin{scope}[x={(image.south east)}, y={(image.north west)}]
            \node[anchor=north west, align=left] at (0, 1) {
                \textbf{Stairs Traversal} \\
                \small Duration: 2.98\,s \\
                \small Avg. Speed: 1.31\,m/s
            };
            \node[anchor=north east, fill opacity=0.0, inner sep=0, xshift=-0pt, yshift=-0pt] at (1, 1) {
                \includegraphics[width=0.26\textwidth]{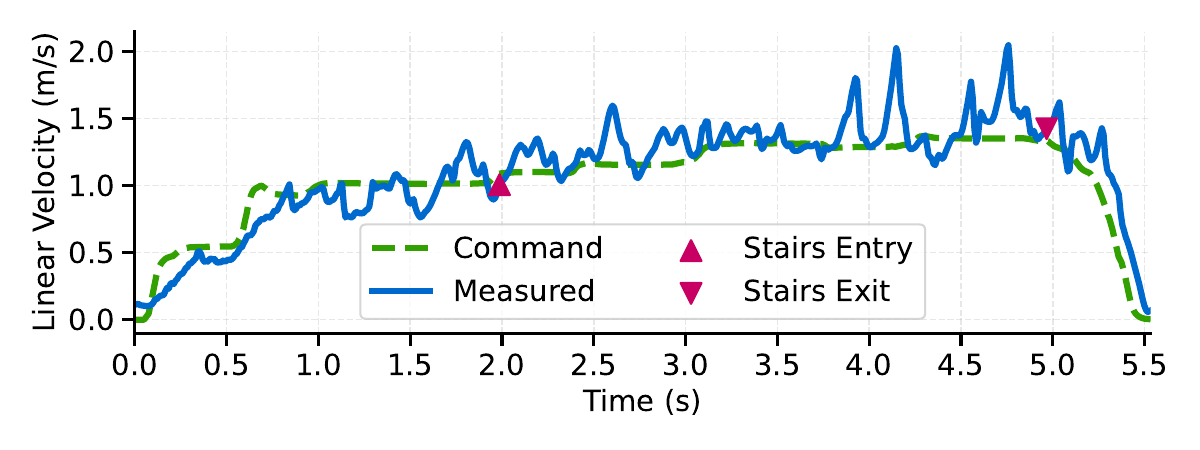}
            };
        \end{scope}
    \end{tikzpicture}
    \caption{Experiment on wooden stairs with a 10 cm rise and 15 cm drop. The upper-right plot depicts the velocity tracking curve captured through a motion capture system where the tracking error is 0.15 m/s.}
    \label{fig:stairs_velocity_tracking}
\end{figure*}

\begin{figure*}[t]
    \centering
    \begin{minipage}{0.646\textwidth}
        \centering
        \begin{tikzpicture}
            \node[anchor=south west, inner sep=0] (image) at (0,0) {\includegraphics[width=\textwidth]{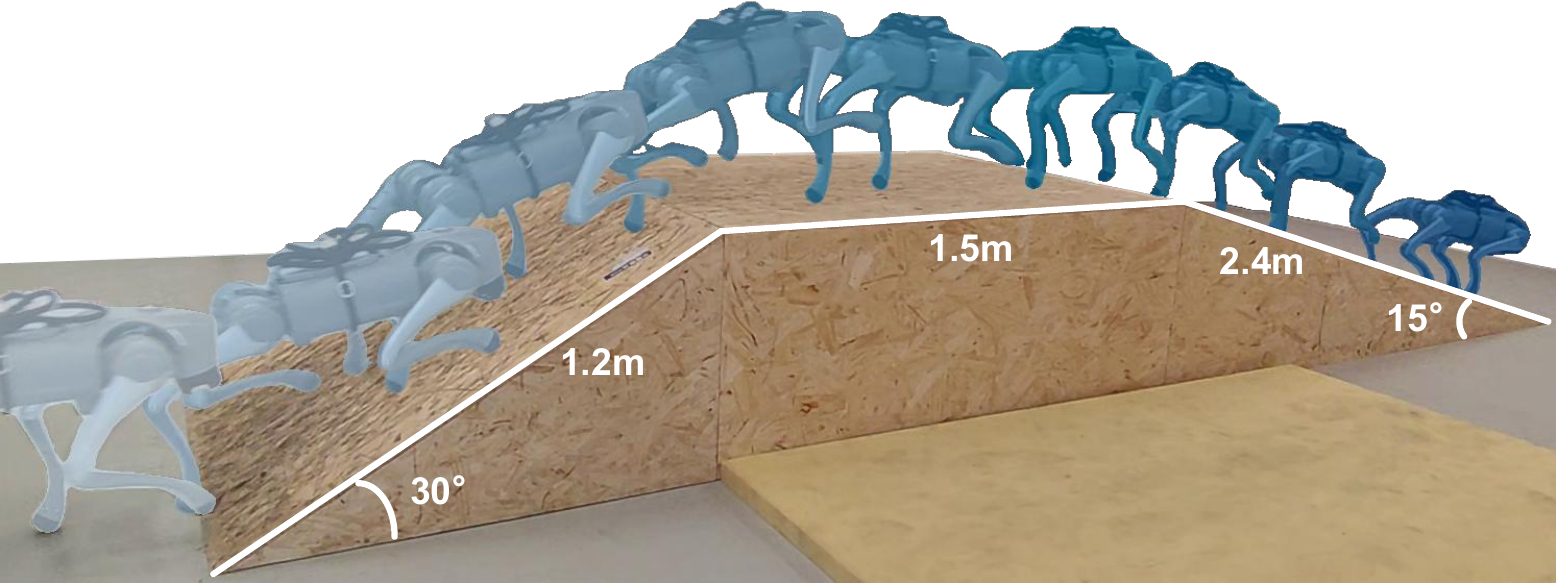}};
            
            \begin{scope}[x={(image.south east)}, y={(image.north west)}]
                \node[anchor=north west, align=left] at (0, 1) {
                    \textbf{Slope Traversal} \\
                    \small Duration: 3.33\,s \\
                    \small Avg. Speed: 1.53\,m/s
                };
            \end{scope}
        \end{tikzpicture}
    \end{minipage}%
    \hfill
    \begin{minipage}{0.348\textwidth}
        \centering
        \begin{tikzpicture}
            \node[anchor=south west, inner sep=0] (image) at (0,0) {\includegraphics[width=\textwidth]{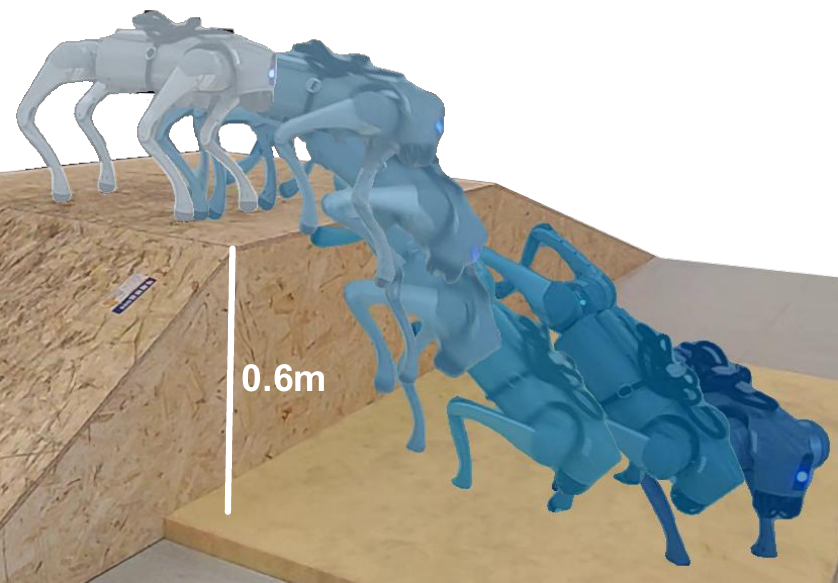}};
            \begin{scope}[x={(image.south east)}, y={(image.north west)}]
                \node[anchor=north east, align=right] at (1, 1) {
                    \textbf{Drop Recovery} \\
                    \small Height: 0.6\,m
                };
            \end{scope}
        \end{tikzpicture}
    \end{minipage}
    
    \caption{Robust locomotion during slope traversal and drop recovery. The left panel highlights a 1.7 s efficiency gain on $\mu=0.71$ slopes compared to the built-in RL baseline and the right frame verifies reliable recovery from 60 cm drops.}
    \label{fig:slope_traveral_and_stability}
    \vspace{-0.50cm}
\end{figure*}

\begin{figure}[htbp]
    \begin{minipage}{\linewidth}
        \centering
        \vspace{-0.1cm}
        \begin{minipage}{0.65\linewidth}
            \centering
            \includegraphics[width=\textwidth]{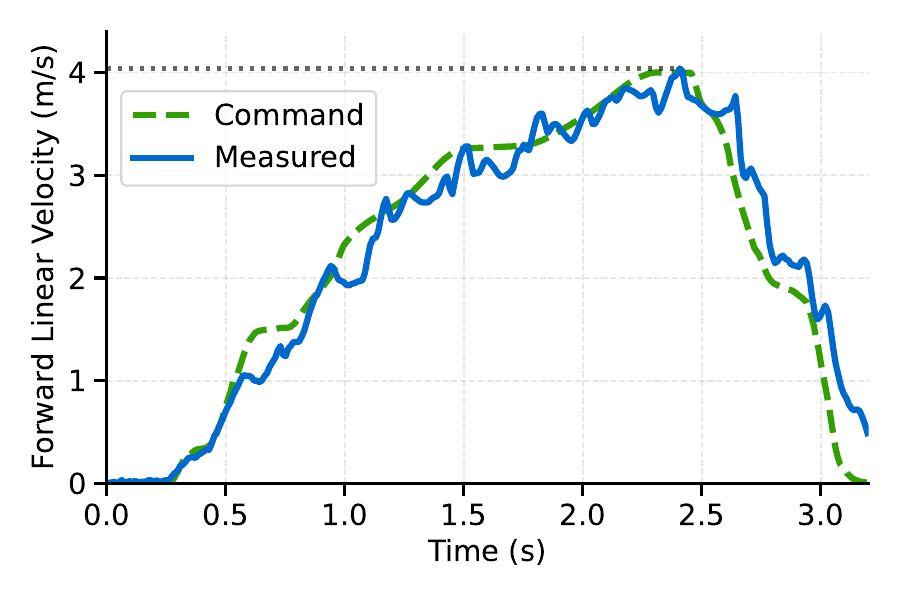}
        \end{minipage}
        \begin{minipage}{0.25\linewidth}
            \begin{minipage}{\linewidth}
                \centering
                \includegraphics[width=\textwidth]{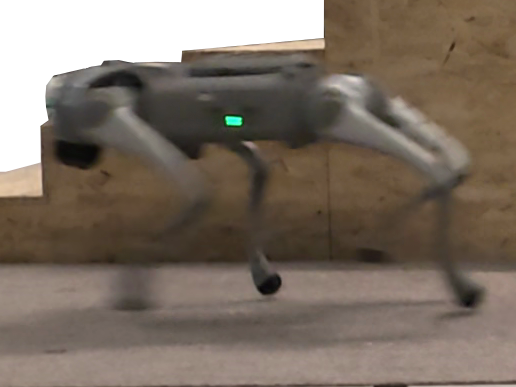}
            \end{minipage}
            \begin{minipage}{\linewidth}
                \centering
                \includegraphics[width=\textwidth]{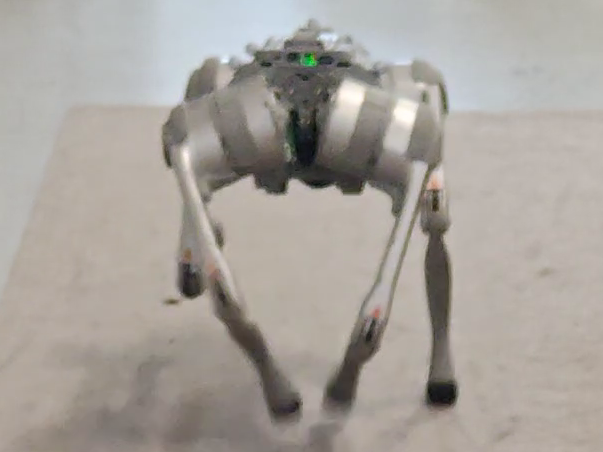}
            \end{minipage}
        \end{minipage}
        \caption{
        Velocity tracking and gait on a $\mu=0.6$ surface. The left plot exhibits command following reaching 4.01 m/s within 2.16 s with a 0.20 m/s error. The upper-right image captures transient flight phases while the lower-right image highlights a stable narrow-base gait.}
        \label{fig:flat_velocity_tracking}
    \end{minipage}

    \vspace{0.3cm}

    \begin{minipage}{\linewidth}
        \centering
        \includegraphics[width=\linewidth]{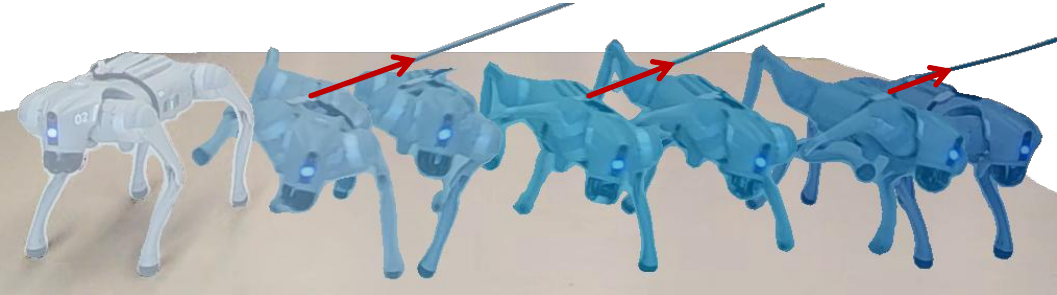}
        \caption{Continuous lateral pull disturbance rejection experiment on flat terrain. The robot endures repeated lateral pulls of approximately $25\sim40$N while maintaining stable locomotion.}
        \label{fig:continuous_lateral_pull}
    \end{minipage}
    \vspace{-0.5cm}
\end{figure}

\begin{table}[!h]
\centering
\caption{Real-World Survival Rate Comparison}
\label{table:survival_rate_optimized}
\newcolumntype{C}{>{\centering\arraybackslash}X}
\begin{tabularx}{\linewidth}{
    l
    >{\centering\arraybackslash\hsize=0.9\hsize}X
    >{\centering\arraybackslash\hsize=1.2\hsize}X
    >{\centering\arraybackslash\hsize=0.9\hsize}X
}
\toprule
\multirow{3}{*}{\textbf{Model}} & \multicolumn{3}{c}{\textbf{Survival Rate (\%)} $\uparrow$} \\
\cmidrule(l){2-4}
& \shortstack{\textbf{Lat. Impulse} \\ (80--100 N)} & \shortstack{\textbf{Tile Stairs 15.5cm} \\ ($\mu=0.38$)} & \shortstack{\textbf{Obstacle 30cm} \\ ($\mu=0.85$)} \\
\midrule
\textbf{Ours} & \textbf{18/20} & \textbf{85/85} & \textbf{17/20} \\
Built-in RL   & 5/20           & \textbf{85/85} & 0/20  \\
CTS           & 11/20          & 18/85          & 0/20  \\
HIM           & 8/20           & 24/85          & 0/20  \\
DreamWaQ      & 7/20           & 12/85          & 0/20  \\
\bottomrule
\end{tabularx}
\vspace{-0.2cm}
\end{table}

\subsection{Comparison on Terrain Challenges}
We deployed the proposed model and baselines on a Unitree Go2 quadruped robot to evaluate its real-world performance as summarized in Table \ref{table:survival_rate_optimized}.
The experimental validation comprises three robustness scenarios including sudden lateral pulls between 80 N and 100 N then 15.5 cm smooth tile stairs and 30 cm obstacle climbing where Appendix Fig. \ref{fig:challenging_scenarios} depicts the specific setups.
Only our model successfully surmounted the 30 cm obstacle while also exhibiting the most effective disturbance rejection during lateral pulls.
Although both our approach and the built-in reinforcement learning controller conquered the stairs, our model completed the 85 steps 17 s faster than the baseline.

\subsection{Velocity Tracking Precision}
We employed a motion capture system to assess velocity tracking accuracy across both flat terrain and stair scenarios. Fig. \ref{fig:stairs_velocity_tracking} depicts the robot traversing stairs at an average speed of 1.31 m/s with a tracking error of 0.15 m/s, which confirms the robust tracking proficiency of the framework even when tackling complex environments.
We further evaluated the locomotion performance on a 30 degree wooden slope where the robot maintains an average velocity of 1.53 m/s. This efficiency reduces the traversal duration by 1.7 s compared to the built-in reinforcement learning baseline as documented in Fig. \ref{fig:slope_traveral_and_stability} of the Appendix.

Fig. \ref{fig:flat_velocity_tracking} illustrates the tracking performance during high-speed locomotion on flat ground. Restricted by an 8 m indoor runway, the robot attains a peak velocity of 4.01 m/s within 2.16 s with a tracking error of 0.20 m/s, which demonstrates exceptional acceleration and braking capabilities. Notably the model autonomously develops a stable narrow-base gait despite the absence of explicit motion constraints to minimize lateral center-of-mass oscillations and bolster stability during high-speed maneuvers.

\subsection{Stability and Generalization}

\begin{figure*}[!p]
    \centering
    \begin{minipage}[t]{0.48\textwidth}
        \vspace{0pt}
        \centering
        \includegraphics[width=\linewidth]{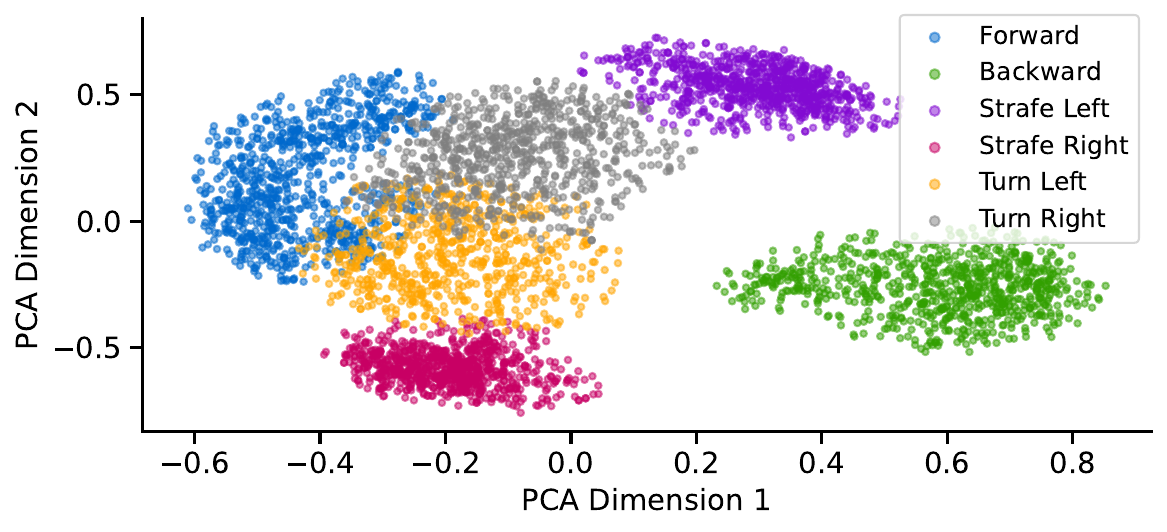}
        \captionof{figure}{PCA visualization of the student encoder latent space in different commands with all terrains.}
        \label{fig:cmd_pca_plot}
    \end{minipage}
    \hfill
    \begin{minipage}[t]{0.48\textwidth}
        \vspace{0pt}
        \centering
        \begin{minipage}{\linewidth}
            \centering
            \includegraphics[width=\textwidth]{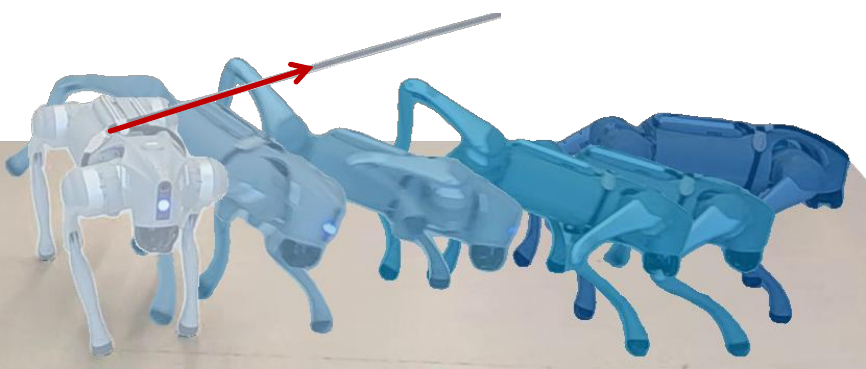}
        \end{minipage}

        \vspace{0.1cm}

        \begin{minipage}{0.50\linewidth}
            \centering
            \includegraphics[width=\textwidth]{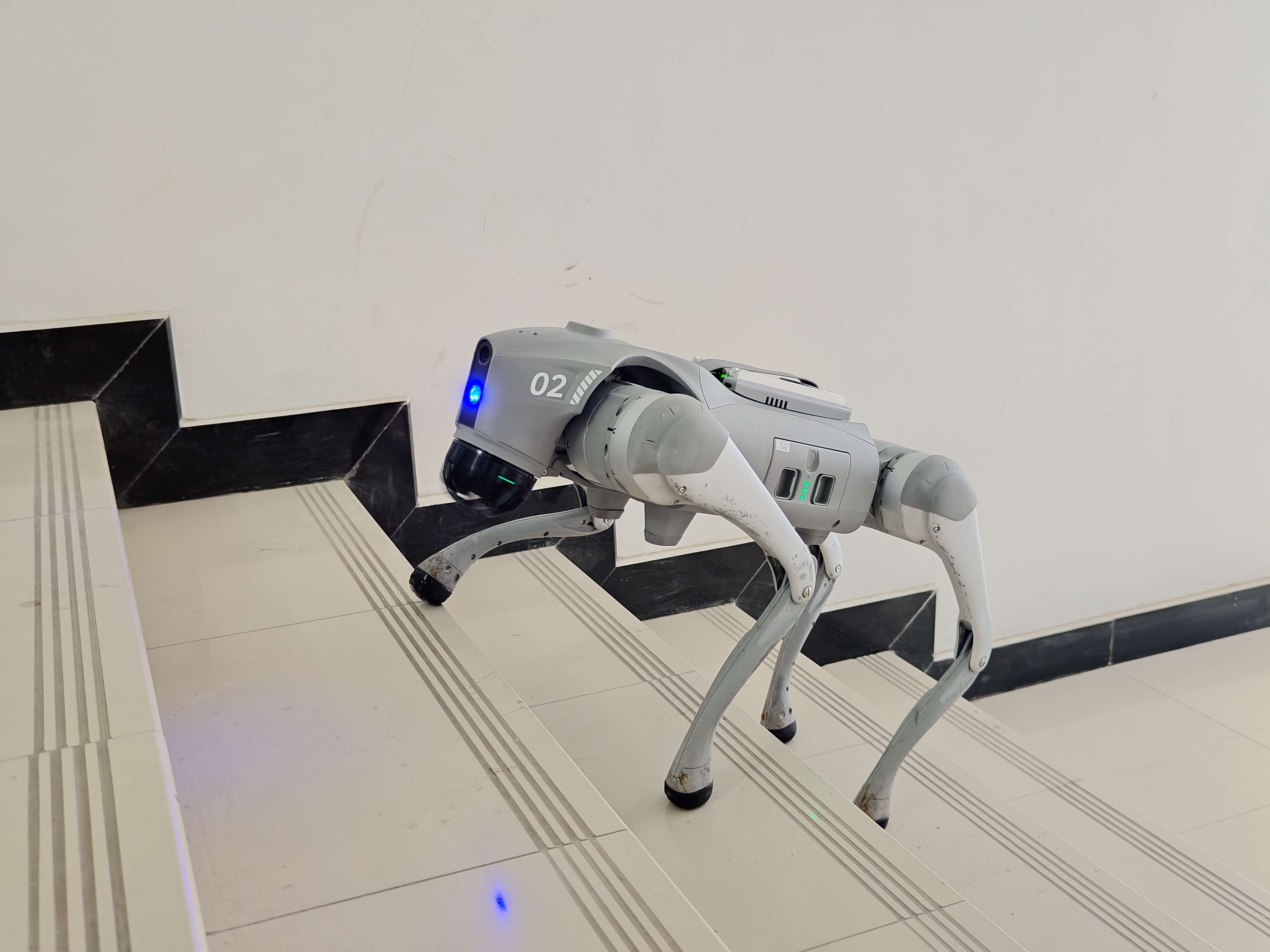}
        \end{minipage}
        \begin{minipage}{0.48\linewidth}
            \centering
            \includegraphics[width=\textwidth]{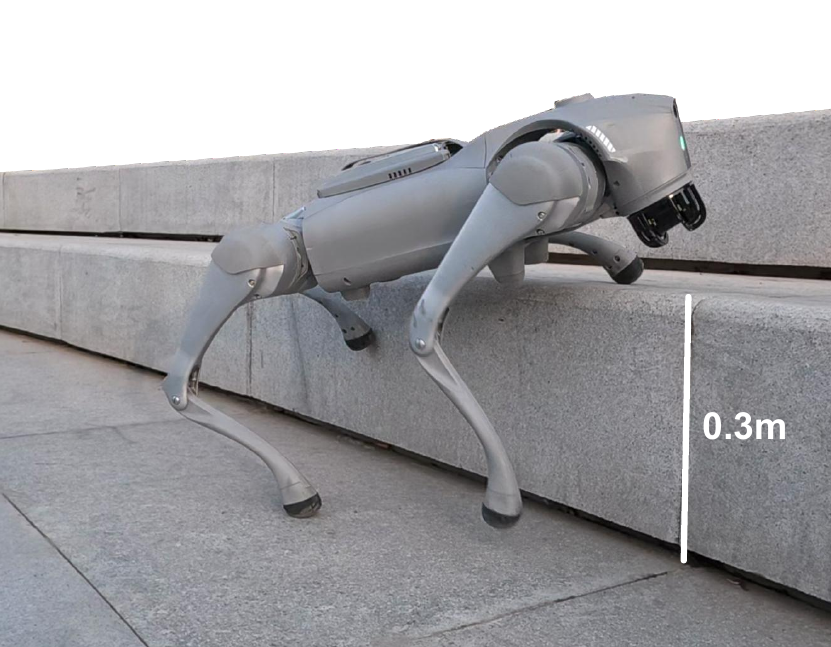}
        \end{minipage}
        \captionof{figure}{Locomotion performance of the Unitree Go2 across three challenging scenarios. Top image illustrates the robot maintaining balance against a lateral impulse between 80 N and 100 N. Bottom-left image depicts the stable ascent of 15.5 cm tile stairs with $\mu=0.38$. Bottom-right image showcases the successful traversal of a 30 cm obstacle where $\mu=0.85$.}
        \label{fig:challenging_scenarios}
    \end{minipage}

    \vspace{0.3cm}

    \begin{minipage}[t]{0.78\textwidth}
        \centering
        \includegraphics[width=0.5\linewidth]{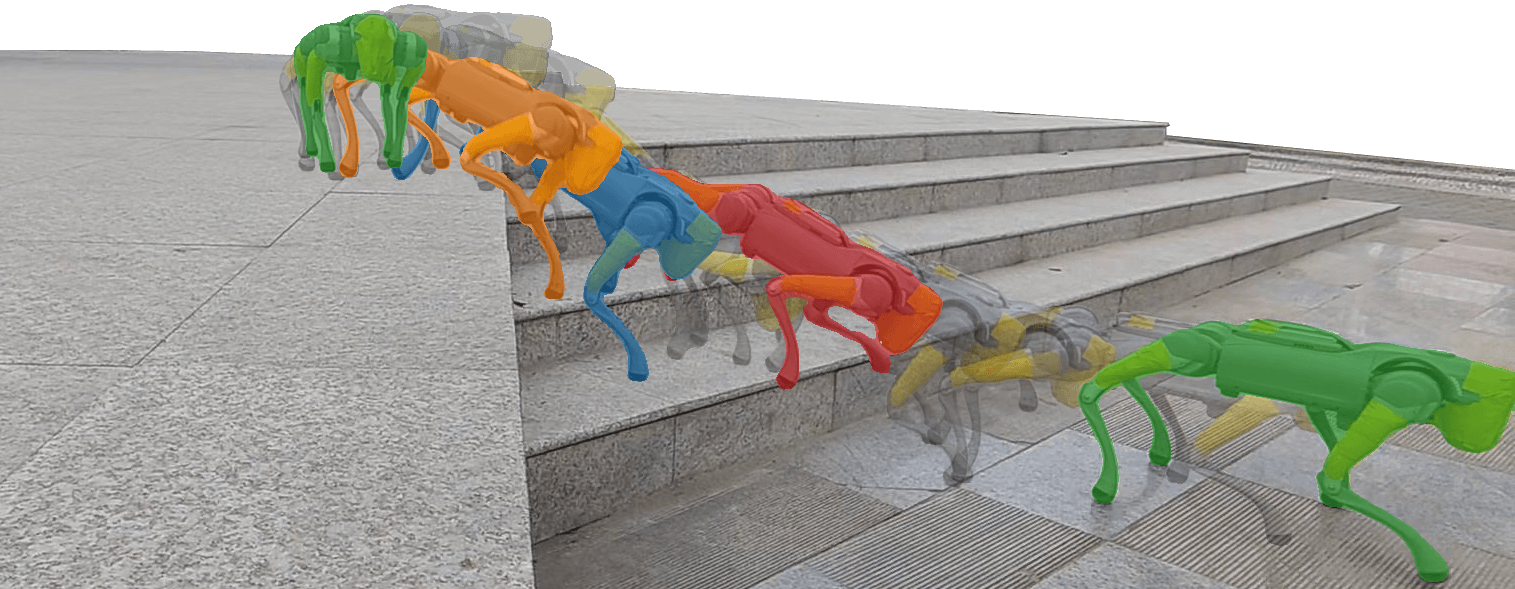}
        \includegraphics[width=\linewidth]{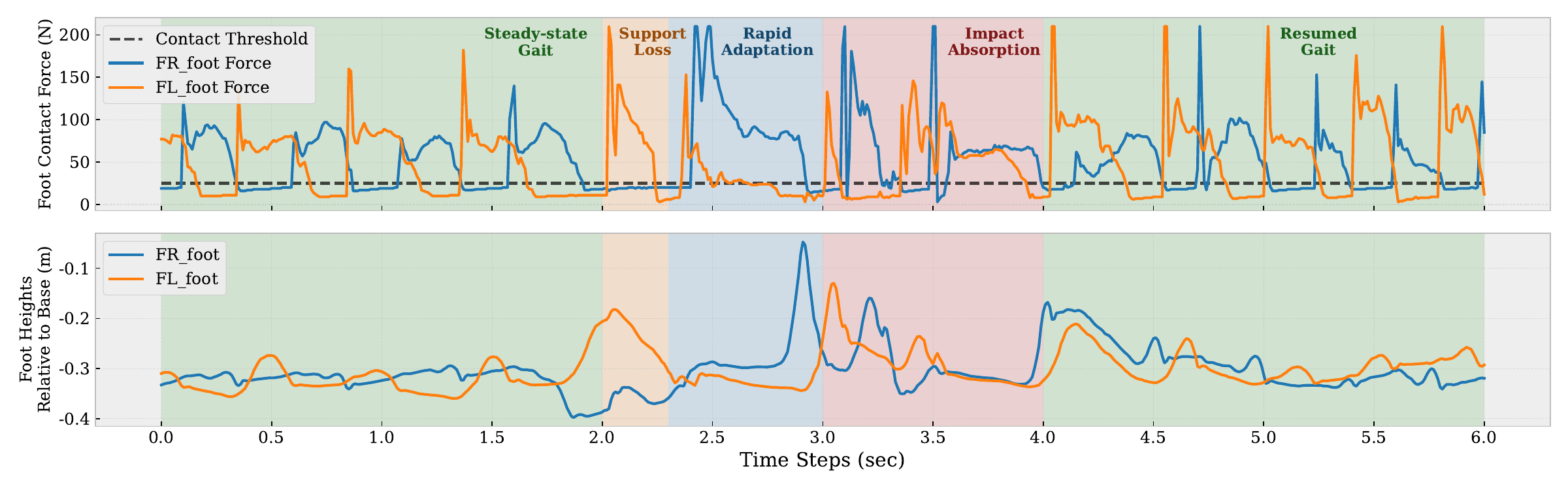}
        \captionof{figure}{The top panel shows the robot quickly adjusting its posture to safely descend when the ground ends at the edge. The middle plot depicts the contact force signals measured by the foot sensors. The bottom image illustrates the front foot height relative to the base calculated from forward kinematics. These results confirm the robustness of the policy and its capacity for adaptive gait transitions across diverse challenges.}
        \label{fig:unexpected_recovery}
    \end{minipage}
\end{figure*}

We validated the emergency recovery capabilities of the proposed model across two challenging real-world scenarios.
First, the robot is subjected to external forces such as strong pushes or pulls where it shows great disturbance rejection by changing its center of mass and creating gaits to offset the impact. Fig.~\ref{fig:continuous_lateral_pull} and \ref{fig:challenging_scenarios} show that the robot remains stable under continuous lateral pulls between 25 N and 40 N as well as sudden impulses of 85 N to 100 N where established baselines almost entirely fail to maintain balance.
Second, when encountering a sudden loss of support the robot rapidly reconfigures its gait to secure its footing and prevent forward tumbling. Fig.~\ref{fig:slope_traveral_and_stability} illustrates a successful recovery sequence from a 60 cm drop while Fig.~\ref{fig:unexpected_recovery} depicts the natural transition to a stable posture after an unexpected fall from flat ground onto stairs.

Finally we conducted field tests in diverse outdoor environments to evaluate the generalization capabilities of the framework. The right panel of Fig. \ref{fig:framework} illustrates the performance across various terrains such as sand and ice as well as slopes and uneven terrains.
The robot completed all trials with a 100\% success rate and zero unexpected terminations which highlights the exceptional robustness of the learned policy.

\section{Conclusions and future work}
In this work, we presented a training framework comprising the RoboGauge assessment suite and an MoE locomotion policy which enables robust multi-terrain locomotion relying solely on proprioception.
Physical experiments on a Unitree Go2 robot demonstrate that our framework successfully surmounts challenging environments including 30 cm obstacles and 100 N impulses, while utilizing the identical training configuration on flat ground to attain a peak velocity of 4.01 m/s.
The framework consistently outperforms established baselines in both tracking precision and recovery stability with a 100\% success rate in diverse outdoor field tests.
This synergy between predictive assessment and modular architecture provides a reliable and efficient way to bridge the gap between simulation results and actual physical performance.

Future research will extend RoboGauge to broader morphologies like humanoid robots and integrate exteroceptive perception with the MoE representation to further improve the crossing of complex structural obstacles.

\section*{Acknowledgement}
The authors thank Tencent AI Arena and Unitree Robotics for providing the Go2 quadruped for the initial experiments. Thanks to Guangsheng Li for providing the training code for baseline DreamWaQ. 

This work was supported in part by the National Key R\&D Program of China under Grant No. 2024YFB4707600, NSFC under grant No. 62125305, No. U23A20339, No. 62573339, No. 62503380, Fundamental and Interdisciplinary Disciplines Breakthrough Plan of the Ministry of Education of China under grant No. JYB2025XDXM210, Natural Science Foundation of Shaanxi Province under Grant 2025SYSSYSZD-083, and State Grid Corporation of China Science and Technology Projects 52060025005B-439-ZN.

\bibliographystyle{unsrtnat}
\bibliography{references}

\clearpage
\appendices

\section{RoboGauge Supplementary Material}
\label{appendix:robogauge}
\subsection{Stability Metric}
\label{appendix:stability_metric}
To provide a more comprehensive evaluation of locomotion stability, we introduce two formal physical criteria in RoboGauge: the Zero Moment Point (ZMP) margin \cite{ZMP2004} and a Coulomb friction margin under Contact Wrench Cone (CWC) constraints \cite{CWC2015}.

\subsubsection{Zero Moment Point (ZMP) Margin}

The Zero Moment Point (ZMP) is a fundamental concept in legged locomotion, defined as the point on the ground where the net moment of inertial and gravitational forces has no horizontal components. To formalize this metric within our framework, we establish the following definitions:
\begin{itemize}
    \item \textbf{Support Polygon:} The convex hull formed by all active contact points between the robot and the ground.
    \item \textbf{Fictitious ZMP (FZMP):} When the calculated ZMP falls outside the support polygon, it is referred to as the FZMP, indicating that the system is in a dynamically unbalanced state.
    \item \textbf{Virtual Horizontal Plane:} A coordinate frame translated to $O'$, representing the geometric center of all active ground contact points, which projects the system onto the $xy$-plane.
\end{itemize}

Within the MuJoCo simulation environment, the ZMP is calculated by aggregating the dynamics over all $N$ rigid bodies of the robot. For the $i$-th rigid body at the current timestep, we define its mass as $m_i$, its center of mass (CoM) position relative to $O'$ as $\boldsymbol{p}_i$, its CoM linear acceleration as $\boldsymbol{\ddot{p}}_i$, its angular velocity as $\boldsymbol{\omega}_i$, its angular acceleration as $\boldsymbol{\dot{\omega}}_i$, and its inertia tensor as $\boldsymbol{I}_i$. It is crucial that all kinematic and inertial properties are strictly expressed in the world coordinate frame.

The total force $\boldsymbol{F}_{\text{total}}$ and total moment $\boldsymbol{M}_{\text{total}}$ of the system are formulated as:
\begin{align}
\boldsymbol{F}_{\text{total}} =&\ \sum_{i=1}^{N} m_i (\boldsymbol{g} - \boldsymbol{\ddot{p}}_i)\\
\boldsymbol{M}_{\text{total}} =&\ \sum_{i=1}^{N} \left[ (\boldsymbol{p}_i \times m_i (\boldsymbol{g} - \boldsymbol{\ddot{p}}_i)) - (\boldsymbol{I}_i \boldsymbol{\dot{\omega}}_i + \boldsymbol{\omega}_i \times (\boldsymbol{I}_i \boldsymbol{\omega}_i)) \right]
\end{align}

By definition, the relationship between the total moment and force at the ZMP is given by $\boldsymbol{M} = \boldsymbol{r}_{\text{zmp}} \times \boldsymbol{F}_{\text{total}}$. Expanding this cross product yields the moment components:
\begin{equation}
\begin{cases} \boldsymbol{M}_y = -x_{\text{zmp}}\boldsymbol{F}_z + z_{\text{zmp}}\boldsymbol{F}_x \\ \boldsymbol{M}_x = y_{\text{zmp}}\boldsymbol{F}_z - z_{\text{zmp}}\boldsymbol{F}_y \end{cases}
\end{equation}

By projecting the ZMP onto the virtual horizontal plane ($z_{\text{zmp}}=0$), the exact ZMP coordinates $(x_{\text{zmp}}, y_{\text{zmp}})$ are derived as:
\begin{equation}
x_{\text{zmp}} = -\frac{\boldsymbol{M}_y}{\boldsymbol{F}_z},\quad y_{\text{zmp}} = \frac{\boldsymbol{M}_x}{\boldsymbol{F}_z}
\end{equation}

Let $D_{\text{norm}}$ denote the diagonal stance span of the robot in its default posture. The normalized ZMP Margin, representing the horizontal distance error of the ZMP relative to the geometric center of the active contacts, is defined as:
\begin{equation}
m_{\text{zmp margin}} = \max\left(0, 1 - \frac{||(x_{\text{zmp}}, y_{\text{zmp}})||_2}{D_{\text{norm}}}\right)
\end{equation}

\subsubsection{Coulomb Friction Margin}

To account for potential slippage and Contact Wrench Cone (CWC) constraints, we introduce a translational friction margin. Let $N_c$ be the number of active foot contacts with the ground. For each contact $i$, $f_i^{\text{tangent}}$ represents the tangential force, $f_i^{\text{normal}}$ represents the normal force, and $\mu$ is the surface friction coefficient.

The Coulomb Friction Margin is calculated as the normal-force-weighted average slack to the friction-cone boundary over all active contacts:
\begin{equation}
m_{\text{friction margin}} = \sum_{i=1}^{N_c} w_i \max\left(0, 1 - \frac{||f_i^{\text{tangent}}||}{\mu f_i^{\text{normal}}}\right) 
\end{equation}

where the weighting factor $w_i$ dynamically emphasizes contacts bearing greater vertical loads:
\begin{equation}
w_i = \frac{f_i^{\text{normal}}}{\sum_{j=1}^{N_c} f_j^{\text{normal}}} 
\end{equation}

\subsection{Hyperparameter Configuration}
In the quality score calculation Eq.~\ref{eq:quality_score}, the metric weights are set to $w_k=2$ for task-completion metrics and $w_k=1$ to others.
For the overlapping scoring function Eq.~\ref{eq:overlapping_scoring_function}, the hyperparameters are set to $\alpha=0.09$ and $\beta=0.19$, which ensures the performance score is bounded within the range $[0, 1]$.
Domain randomizations include friction coefficients from $0.2$ to $1.0$ in increments of $0.1$. Terrain levels are designed with difficulty parameters $d$ ranging from $0.1$ to $1.0$ in increments of $0.1$ as detailed in Table~\ref{tab:terrain_design}. The locomotion control objectives are configured as described in Table~\ref{table:goals}.

\subsection{Implementation Details}

The operational logic for each pipeline is delineated below.

The \texttt{BasePipeline} (Fig. \ref{fig:base_pipeline}) orchestrates the interaction between the simulation engine \texttt{sim}, the evaluator \texttt{gauge} responsible for control commands and metric computation, and the locomotion model \texttt{robot}. Additionally, it manages exception handling, domain randomization, and the application of observation noise.

The \texttt{MultiPipeline} leverages multiprocessing to execute the \texttt{BasePipeline} across diverse seeds and domain randomization configurations while aggregating the output files. To determine the maximum navigable difficulty for a given terrain, the \texttt{LevelPipeline} (Fig. \ref{fig:level_pipeline}) identifies the highest level that the model traverses successfully across three separate random seeds. 

\newpage
\begin{strip}
\centering
\captionof{table}{Terrain Level Design Parameters ($d=0.1,0.2,\cdots,1.0$)}
\label{tab:terrain_design}
\begin{tabularx}{\textwidth}{
    >{\raggedright\arraybackslash\hsize=.5\hsize}X 
    >{\raggedright\arraybackslash\hsize=1.6\hsize}X 
    >{\centering\arraybackslash\hsize=0.9\hsize}X
}
\toprule
\textbf{Terrain} & \textbf{Specification \& Formula} & \textbf{Range / Unit} \\ 
\midrule
\multirow{3}{*}{\textbf{Wave}} & Amplitude $A = 0.4d$ & $A\in [0.04, 0.4]$ \\
    & Period $T = 1.6$ & m \\
    & Height Field $z(x,y) = A\sin(x/T) + A\cos(y/T)$ & m \\ 
\midrule
\textbf{Slope} & Slope $k = 0.07 + 0.5d$ & Angle $\theta \in [8.4^\circ, 29.7^\circ]$ \\ 
\midrule
\multirow{3}{*}{\textbf{Stairs}} & Step Width $w = 0.31$ & m \\
    & Step Height $h = \begin{cases} 
        0.05+0.3d, & 0.1 \le d \le 0.4 \\
        0.17+0.1(d-0.4), & 0.5 \le d \le 1.0 
    \end{cases}$ & $h \in [0.08, 0.23]$ m \\ 
\midrule
\textbf{Obstacles} & Height $h = 0.05 + 0.23d$ & $h \in [0.073, 0.28]$ m \\ 
\bottomrule
\end{tabularx}

\vspace{0.5cm}

\captionof{table}{Configuration of Locomotion Control Objectives}
\label{table:goals} 
\begin{tabularx}{\textwidth}{
    >{\raggedright\arraybackslash\hsize=.6\hsize}X 
    >{\raggedright\arraybackslash\hsize=1.2\hsize}X 
    >{\raggedright\arraybackslash\hsize=.7\hsize}X 
    >{\centering\arraybackslash\hsize=.3\hsize}X
}
\toprule
\textbf{Goal Name} & \textbf{Description} & \textbf{Reset Condition} & \textbf{Max Trials} \\ \midrule
Max Velocity & Evaluation of peak linear or angular velocity in a single dimension. & Sudden stop after each directional command. & 6 \\
Diagonal Velocity & Tracking of coupled diagonal velocity vectors (combined linear and angular). & Completion of each pair of diagonal commands. & 8 \\
Target Pos. Velocity & Position-based tracking using a Proportional controller to reach targets. & Goal reached or time limit exceeded. & 1 \\ \bottomrule
\end{tabularx}

\vspace{0.5cm}

\centering
\includegraphics[width=\linewidth]{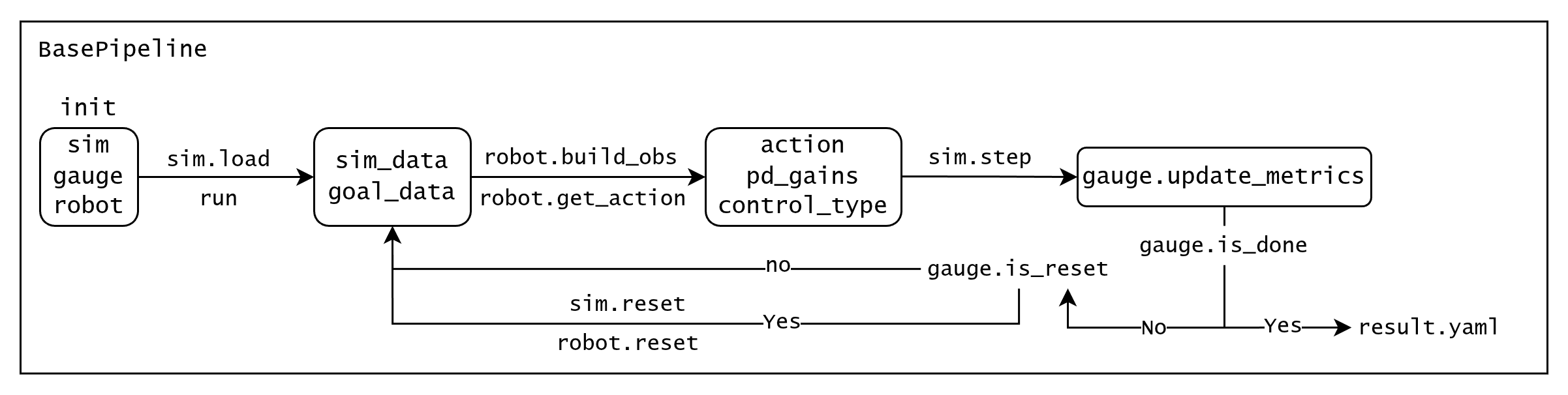}
\captionof{figure}{Operational workflow of the \texttt{BasePipeline}.}
\label{fig:base_pipeline}

\vspace{0.5cm}

\centering
\includegraphics[width=\linewidth]{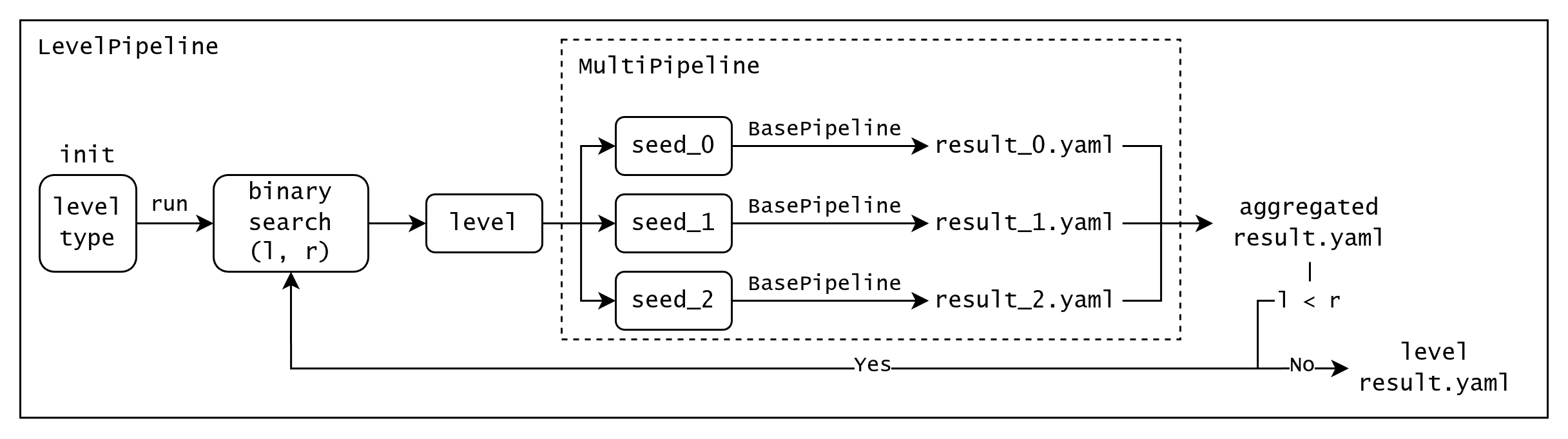}
\captionof{figure}{Operational workflow of the \texttt{LevelPipeline}.}
\label{fig:level_pipeline}

\end{strip}

\section{Training Details}
\label{appendix:training_details}

\subsection{Dynamic Velocity Tracking Precision Adjustment}
\label{appendix:dynamic_sigma}

To adapt the velocity tracking precision $\sigma$ according to terrain characteristics and difficulty levels,
we implement a dynamic scaling adjustment.
We observe that as the maximum command range expands from 0.5 to 1.5,
locomotion on challenging terrains such as wave, stairs,
and obstacle often fails to accurately track the commanded linear velocity.
Consequently, we scale the tracking coefficients to relax the tracking constraints for these scenarios. 

We define $[v_{min}, v_{max}]$ as the velocity magnitude range designated for the dynamic adjustment of $\sigma$.
The parameter $\sigma_{\text{max}}^{T_i}$ denotes the maximum velocity tracking coefficient
assigned to the $i$-th terrain type.
Given a commanded velocity $v$ for the $i$-th terrain,
the intermediate coefficient $\sigma_{\text{vel}}$ is formulated as follows:
\begin{equation}
\begin{cases}
\sigma,&v\in[0,v_{min}),\\
\sigma(v-v_{min})+\sigma_{\text{max}}^{T_i}(v_{max}-v),&v\in[v_{min},v_{max}),\\
\sigma_{\text{max}}^{T_i},&v\in[v_{max},\infty).
\end{cases}
\end{equation}

The final adaptive tracking coefficient $\sigma_{\text{now}}$ incorporates the terrain difficulty level $L$ as delineated below:
\begin{equation}
\sigma_{\text{now}}=\sigma+\min(e^{\frac{L}{10}}-1,1)(\sigma_{\text{vel}}-\sigma)
\end{equation}
The velocity commands $v$ pertain to longitudinal and lateral linear velocities
as well as angular velocity commands.
Table \ref{table:sigma_cmd_limit} in the Appendix details the maximum velocity tracking coefficients
$\sigma_{\text{max}}^{T_i}$ and the associated velocity adjustment ranges across diverse terrains.

\subsection{Command Design}
\label{appendix:command_design}
Direct training with the full command range of $[-1, 1]$ m/s
across all terrains enables rapid progression through difficulty levels
but frequently yields unstable gaits.
Specifically, the robot often demonstrates erratic behaviors such as leaping and high-frequency leg motions.
Conversely, training from low-speed commands facilitates
the acquisition of stable locomotion patterns.
We therefore introduce a \textit{command curriculum} to address these issues,
as detailed in Table~\ref{table:cmd_curriculum}.

We observed that when the maximum command magnitude exceeds $[-1, 1]$ m/s,
the robot fails to accurately track the target linear velocity on complex terrains such as wave,
stairs, and obstacle.
This tracking discrepancy induces instability during the training process.
Therefore, we impose specific constraints on the maximum command range for
individual terrains as detailed in Table~\ref{table:sigma_cmd_limit}.
Notably, although these limits are strictly enforced during the training phase,
no such restrictions are applied during hardware testing on the physical robot.
Despite this discrepancy, the model follows commands that lie beyond
the training distribution and demonstrates robust generalization capabilities.

Our empirical analysis indicates that uniform sampling distributions are suboptimal
because boundary values exhibit an exceptionally low probability of occurrence
despite being frequently encountered during hardware deployment.
To address this issue, we introduced an \textit{extreme command sampling} strategy.
This methodology allocates a 10\% probability to stationary commands and
a 20\% probability to command combinations
that represent maximum velocity limits across all three dimensions.
Furthermore, when the linear velocity is zero,
the framework maintains a 20\% probability of sampling the maximum angular velocity
to enhance robustness during pivot turns.

At the start of training,
the linear velocity command range is restricted to $[-0.5, 0.5]$ m/s with
a 10\% probability of remaining stationary.
Such a narrow distribution frequently produces command sequences that fail the terrain level-up condition,
which necessitates a final horizontal distance relative to the initial position exceeding 4m, a value equivalent to half the terrain length \cite{rudin2022learning}.
This limitation prevents the agent from exploring higher difficulty levels.
To guarantee that the cumulative command length surpasses the required threshold,
we implement a \textit{dynamic command sampling} strategy.

Let $n_r$ represent the number of sampled commands and $\boldsymbol{v}_i^{\text{cmd}}$
denote the $i$-th linear velocity command.
Given that $T_r$ signifies the sampling interval and $T_{\text{ep}}$ is the episode duration,
the sampling range for the $(n_r+1)$-th command is restricted to the intervals
between $(v^{\text{min}},-v^*)\cup(v^*,v^{\text{max}})$ where $v^*$ is formulated as follows:
\begin{equation}
    v^* := \text{clip}\left(\frac{5-||\sum_{i=1}^{n_r}\boldsymbol{v}_i^{\text{cmd}}||_2T_r}{T_{\text{ep}}-n_rT_r},0,\min(|v^{\text{min}}|,|v^{\text{max}}|)\right)
\end{equation}

Should a stationary command be selected for the $(n_r+1)$-th sample,
its specific duration is determined as follows:
\begin{equation}
   T^{zero} = \text{clip}\left(T_{\text{ep}}-n_rT_r-\frac{5-||\sum_{i=1}^{n_r}\boldsymbol{v}_i^{\text{cmd}}||_2T_r}{0.8\times\max(v^{\text{max}}_x,v^{\text{max}}_y)},0,T_r\right) 
\end{equation}

The integration of the aforementioned command curriculum, extreme command sampling,
and dynamic command sampling promotes the development of more stable locomotion gaits
while ensuring a steady advancement across terrain difficulty levels.
Additionally, these strategies markedly raise the performance ceiling for models
evaluated with the RoboGauge.

\begin{figure}[htbp]
    \centering
    \includegraphics[width=\linewidth]{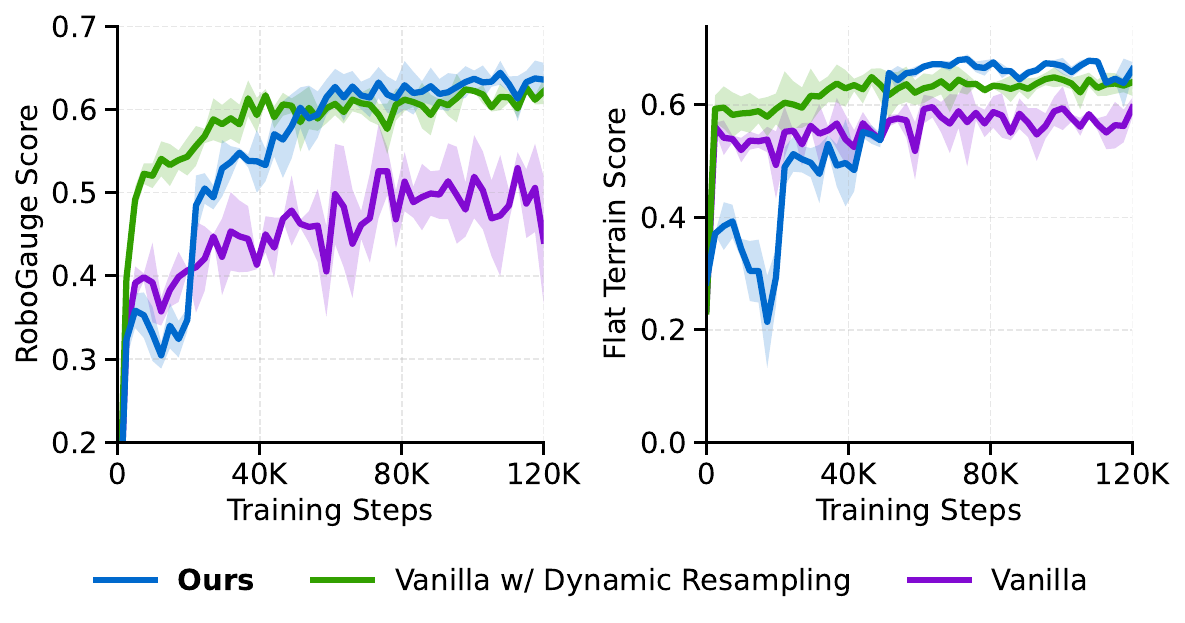}
    \caption{Ablation study on training strategies.}
    \label{fig:training_ablation}
\end{figure}

We conducted ablation studies on the training configurations where Fig. \ref{fig:training_ablation} illustrates the impact of dynamic command sampling. Activating this feature accelerates convergence and elevates the peak reward by 11\% relative to the version without dynamic sampling.
The final training curve is achieved by further incorporating dynamic velocity tracking precision adjustment and a command curriculum. These additions significantly bolster training stability and improve performance on flat terrain.

\section{Train configuration}

\begin{table}[!h]
\centering
\caption{Reward Function Specifications} \label{table:reward}
\begin{tabularx}{\linewidth} {
   >{\arraybackslash}X 
   >{\arraybackslash\hsize=1.2\hsize}X
   >{\arraybackslash\hsize=.6\hsize}X
  }
\toprule
Reward Term & Equation & Weight \\
\midrule
Lin. velocity tracking & $\exp(-\sigma||\boldsymbol{v}_{xy}^{\text{cmd}}-\boldsymbol{v}_{xy}||_2^2)$ & $1.0/\textcolor{red}{2.0}$ \\
Ang. velocity tracking & $\exp(-\sigma|\omega_z^{\text{cmd}}-\omega_z|^2)$ & $0.5$ \\
Lin. velocity ($z$) & $v_z^2$ & $-2.0$ \\
Ang. velocity ($xy$) & $||\boldsymbol{\omega}_{xy}||_2^2$ & $-0.05$ \\
Joint acceleration & $\ddot{q}^2$ & $-2.5\times 10^{-7}$ \\
Joint power & $|\boldsymbol{\tau}||\dot{q}|^T$ & $-2\times 10^{-5}$ \\
Joint torque & $||\boldsymbol{\tau}||_2^2$ & $-1\times 10^{-4}$ \\
Base height & $(h^{\text{des}}-h)^2$ & $-1.0$ \\
Action rate & $||\boldsymbol{a}_t-\boldsymbol{a}_{t-1}||_2^2$ & $-0.01$ \\
Action smoothness & $||\boldsymbol{a}_t-2\boldsymbol{a}_{t-1}+\boldsymbol{a}_{t-2}||_2^2$ & $-0.01$ \\
Collision & $n_{\text{collision}}$ & $-1.0$ \\
Joint limit & $n_{\text{limitation}}$ & $-2.0$ \\
Foot regulation & $r^{\text{fr}}$ & $-0.05$ \\
Hip regulation & $|\boldsymbol{q}^{\text{hip}}-\boldsymbol{q}_{\text{default}}^{\text{hip}}|$ & $-0.05$ \\
\textcolor{red}{Hip symmetry} & \textcolor{red}{$r^{\text{hs}}$} & \textcolor{red}{$-1$} \\
\bottomrule
\end{tabularx}
\begin{tablenotes}
    \footnotesize
    \item Black: Reward terms utilized for the multi-terrain model.
    \item {\color{red}Red: Flat-ground high-speed model modified weights.}
\end{tablenotes}
\end{table}

\begin{table}[htbp]
\centering
\caption{Maximum Velocity Tracking Coefficients and Command Limits Across Terrains}
\label{table:sigma_cmd_limit}
\small
\begin{tabularx}{\linewidth}{l c c c c}
\toprule
\textbf{Terrain Type} & $\sigma_{\text{max}}^{i}$ & $v_x$ [m/s] & $v_y$ [m/s] & $\omega_z$ [rad/s] \\
\midrule
Flat         & 1/4  & $\pm$ 2.0 & $\pm$ 1.0 & $\pm$ 2.0 \\
Wave         & 5/12 & $\pm$ 1.5 & $\pm$ 1.0 & $\pm$ 1.5 \\
Slope        & 1/4  & $\pm$ 1.5 & $\pm$ 1.0 & $\pm$ 1.5 \\
Stairs Up    & 1/2  & $\pm$ 1.0 & $\pm$ 1.0 & $\pm$ 1.5 \\
Stairs Down  & 1/2  & $\pm$ 1.0 & $\pm$ 1.0 & $\pm$ 1.5 \\
Obstacle     & 3/4  & $\pm$ 1.0 & $\pm$ 1.0 & $\pm$ 1.5 \\
\bottomrule
\end{tabularx}
\begin{tablenotes}
    \footnotesize
    \item \textbf{Note:} Velocity ranges are defined as $v^{\text{lin}} \in [0.5, 1.5]$ m/s and $v^{\text{ang}} \in [1.0, 2.0]$ rad/s.
\end{tablenotes}
\end{table}

\begin{table}[htbp]
\centering
\caption{Command Curriculum Stages and Velocity Limits}
\label{table:cmd_curriculum}
\small
\begin{tabularx}{\linewidth}{
    >{\raggedright\arraybackslash\hsize=0.9\hsize}X 
    >{\raggedright\arraybackslash\hsize=1.2\hsize}X 
    >{\centering\arraybackslash\hsize=0.5\hsize}X 
    >{\centering\arraybackslash\hsize=0.5\hsize}X 
    >{\centering\arraybackslash\hsize=0.5\hsize}X 
}
\toprule
\multirow{2}{*}{\textbf{Stage}} & \multirow{2}{*}{\textbf{Training Steps}} & $v_x$ & $v_y$ & $\omega_z$ \\
               &                         & [m/s] & [m/s] & [rad/s] \\
\midrule
Initial      & $[0, 2 \times 10^4]$             & $\pm$ 0.5 & $\pm$ 0.5 & $\pm$ 1.0 \\
Intermediate & $[2 \times 10^4, 5 \times 10^4]$ & $\pm$ 1.0 & $\pm$ 1.0 & $\pm$ 1.5 \\
Advanced     & $[5 \times 10^4, \infty]$       & $\pm$ 2.0 & $\pm$ 1.0 & $\pm$ 2.0 \\
\bottomrule
\end{tabularx}
\end{table}

\section{Supplementary Experiment}

\begin{table*}[!t]
\centering
\footnotesize
\setlength{\tabcolsep}{5pt}
\caption{Comprehensive Evaluation: Real-World Measurements, Predicted Values, and Absolute Errors}
\label{table:full_comparison_data}
\renewcommand{\arraystretch}{1.0}
\resizebox{\textwidth}{!}{%
\begin{tabular}{l l c c c c c c}
\toprule
\textbf{Source} & \textbf{Movement} & \textbf{Lin. Trk.} & \textbf{Ang. Trk.} & \textbf{DOF Power} & \textbf{DOF Limits} & \textbf{Orient.} & \textbf{Smooth.} \\
\midrule
\multirow{4}{*}{\shortstack{Real \\ (Ground Truth)}} & Linear ($x=1$) & 0.9185 & 0.5808 & 0.8527 & 0.9159 & 0.9675 & 0.7739 \\
                               & Lateral ($y=0.5$) & 0.9552 & 0.7037 & 0.9696 & 0.9384 & 0.9661 & 0.8985 \\
                               & Angular ($z=1$) & 0.9552 & 0.8010 & 0.9659 & 0.9439 & 0.9614 & 0.9020 \\
                               & Stairs ($x=1$) & 0.8554 & 0.2732 & 0.6721 & 0.8395 & 0.8749 & 0.5944 \\
\midrule
\multirow{4}{*}{\shortstack{RoboGauge \\ (Predicted)}} & Linear ($x=1$) & 0.8217 & 0.5669 & 0.8330 & 0.9386 & 0.9592 & 0.7853 \\
                                                       & Lateral ($y=0.5$) & 0.8685 & 0.6822 & 0.9704 & 0.9293 & 0.9627 & 0.8861 \\
                                                       & Angular ($z=1$) & 0.8763 & 0.7679 & 0.9734 & 0.9414 & 0.9647 & 0.8983 \\
                                                       & Stairs ($x=1$) & 0.7596 & 0.2507 & 0.6211 & 0.8472 & 0.8625 & 0.6606 \\
\cmidrule(lr){1-8}
\multirow{2}{*}{\shortstack{\textbf{RoboGauge (Ours)} \\ \textbf{(Abs. Error) $\downarrow$}}} & $x=1$ (Merge) & 0.0963 & 0.0182 & 0.0354 & 0.0152 & 0.0103 & 0.0388 \\
                                                                                       & \textbf{Average} & 0.0873 & \textbf{0.0243} & \textbf{0.0145} & \textbf{0.0089} & \textbf{0.0057} & \textbf{0.0183} \\
\midrule
\multirow{4}{*}{\shortstack{IsaacGym \\ (Predicted)}} & Linear ($x=1$) & 0.8977 & 0.7826 & 0.9155 & 0.9361 & 0.9737 & 0.8289 \\
                                                      & Lateral ($y=0.5$) & 0.9694 & 0.8039 & 0.9598 & 0.9378 & 0.9707 & 0.8781 \\
                                                      & Angular ($z=1$) & 0.9853 & 0.9134 & 0.9751 & 0.9325 & 0.9798 & 0.9510 \\
                                                      & Stairs ($x=1$) & 0.8786 & 0.5732 & 0.8635 & 0.9027 & 0.9339 & 0.7454 \\
\cmidrule(lr){1-8}
\multirow{2}{*}{\shortstack{\textbf{IsaacGym} \\ \textbf{(Abs. Error) $\downarrow$}}} & $x=1$ (Merge) & 0.0220 & 0.2509 & 0.1271 & 0.0417 & 0.0326 & 0.1030 \\
                                                                    & \textbf{Average} & \textbf{0.0221} & 0.1545 & 0.0487 & 0.0179 & 0.0185 & 0.0575 \\
\bottomrule
\end{tabular}
}
\end{table*}

\begin{table*}[!t]
\centering
\footnotesize
\setlength{\tabcolsep}{5pt}
\caption{RoboGauge detailed metrics for baselines} \label{table:detailed-baseline-metrics}
\renewcommand{\arraystretch}{1.0}
\resizebox{\textwidth}{!}{%
\begin{tabular}{lcccccccccccc}
\toprule
\multicolumn{1}{c}{\multirow{2}{*}{Model}} & \multicolumn{3}{c}{ang vel err}                     & \multicolumn{3}{c}{lin vel err}                     & \multicolumn{3}{c}{dof limits}                      & \multicolumn{3}{c}{dof power}                       \\
\multicolumn{1}{c}{}                       & mean            & mean@25         & mean@50         & mean            & mean@25         & mean@50         & mean            & mean@25         & mean@50         & mean            & mean@25         & mean@50         \\
\midrule
Our                                        & \textbf{0.7018} & \textbf{0.6047} & \textbf{0.6431} & \textbf{0.7394} & \textbf{0.6497} & \textbf{0.6908} & \textbf{0.8139} & \textbf{0.8029} & \textbf{0.8073} & \textbf{0.7889} & \textbf{0.7411} & \textbf{0.7642} \\
CTS                                        & 0.6231          & 0.5253          & 0.5632          & 0.6464          & 0.5363          & 0.5878          & 0.7341          & 0.7232          & 0.7275          & 0.7113          & 0.6607          & 0.6856          \\
HIM                                        & 0.5652          & 0.4620          & 0.5025          & 0.6297          & 0.5443          & 0.5881          & 0.6781          & 0.6645          & 0.6699          & 0.6529          & 0.5996          & 0.6253          \\
DreamWaQ                                   & 0.5309          & 0.4305          & 0.4698          & 0.5937          & 0.5060          & 0.5512          & 0.6437          & 0.6307          & 0.6360          & 0.6200          & 0.5690          & 0.5939          \\
\bottomrule
\end{tabular}
}

\vspace{0.3cm}

\resizebox{\textwidth}{!}{%
\begin{tabular}{lcccccccccccc}
\toprule
\multicolumn{1}{c}{\multirow{2}{*}{Model}} & \multicolumn{3}{c}{orientation stability}           & \multicolumn{3}{c}{torque smoothness}               & \multicolumn{3}{c}{zmp margin}                      & \multicolumn{3}{c}{friction margin}                 \\
\multicolumn{1}{c}{}                       & mean            & mean@25         & mean@50         & mean            & mean@25         & mean@50         & mean            & mean@25         & mean@50         & mean            & mean@25         & mean@50         \\
\midrule
Our                                        & \textbf{0.8147} & \textbf{0.7946} & \textbf{0.8040} & \textbf{0.7734} & \textbf{0.7400} & \textbf{0.7535} & \textbf{0.7933} & \textbf{0.7401} & \textbf{0.7653} & \textbf{0.6892} & \textbf{0.6051} & \textbf{0.6340} \\
CTS                                        & 0.7346          & 0.7124          & 0.7232          & 0.6954          & 0.6594          & 0.6744          & 0.7163          & 0.6670          & 0.6900          & 0.6188          & 0.5310          & 0.5622          \\
HIM                                        & 0.6780          & 0.6497          & 0.6643          & 0.6416          & 0.6053          & 0.6201          & 0.6604          & 0.6092          & 0.6344          & 0.5544          & 0.4706          & 0.5014          \\
DreamWaQ                                   & 0.6443          & 0.6170          & 0.6312          & 0.6060          & 0.5709          & 0.5853          & 0.6290          & 0.5814          & 0.6049          & 0.5236          & 0.4392          & 0.4706          \\
\bottomrule
\end{tabular}
}
\end{table*}

\begin{table*}[!t]
\centering
\footnotesize
\setlength{\tabcolsep}{5pt}
\caption{RoboGauge detailed terrain scores for baselines} \label{table:detailed-baseline-terrains}
\renewcommand{\arraystretch}{1.0}
\resizebox{\textwidth}{!}{%
\begin{tabular}{lcccccccccccc}
\toprule
\multicolumn{1}{c}{\multirow{2}{*}{Model}} & \multicolumn{3}{c}{flat}                            & \multicolumn{3}{c}{wave}                            & \multicolumn{3}{c}{slope forward}                   & \multicolumn{3}{c}{slope backward}                  \\
\multicolumn{1}{c}{}                       & mean            & mean@25         & mean@50         & mean            & mean@25         & mean@50         & mean            & mean@25         & mean@50         & mean            & mean@25         & mean@50         \\
\midrule
Our                                        & $\mathbf{0.755 \pm 0.005}$     & $\mathbf{0.559 \pm 0.013}$ & $\mathbf{0.663 \pm 0.009}$ & $\mathbf{0.615 \pm 0.042}$ & $\mathbf{0.542 \pm 0.042}$ & $\mathbf{0.579 \pm 0.042}$ & $\mathbf{0.583 \pm 0.104}$ & $\mathbf{0.519 \pm 0.094}$ & $\mathbf{0.551 \pm 0.099}$ & $0.584 \pm 0.105$          & $0.522 \pm 0.094$          & $0.553 \pm 0.100$          \\
CTS                                        & $0.721 \pm 0.004$          & $0.436 \pm 0.008$          & $0.592 \pm 0.006$          & $0.602 \pm 0.056$          & $0.525 \pm 0.055$          & $0.561 \pm 0.055$          & $0.557 \pm 0.104$          & $0.488 \pm 0.094$          & $0.521 \pm 0.098$          & $0.524 \pm 0.101$          & $0.458 \pm 0.090$          & $0.491 \pm 0.096$          \\
HIM                                        & $0.739 \pm 0.003$          & $0.486 \pm 0.009$          & $0.621 \pm 0.006$          & $0.569 \pm 0.030$          & $0.490 \pm 0.028$          & $0.525 \pm 0.029$          & $0.492 \pm 0.090$          & $0.425 \pm 0.082$          & $0.457 \pm 0.085$          & $0.571 \pm 0.102$          & $0.504 \pm 0.092$          & $0.537 \pm 0.097$          \\
DreamWaQ                                   & $0.736 \pm 0.003$          & $0.485 \pm 0.010$          & $0.620 \pm 0.006$          & $0.532 \pm 0.042$          & $0.458 \pm 0.041$          & $0.491 \pm 0.041$          & $0.404 \pm 0.083$          & $0.343 \pm 0.069$          & $0.373 \pm 0.076$          & $\mathbf{0.595 \pm 0.105}$ & $\mathbf{0.533 \pm 0.097}$ & $\mathbf{0.564 \pm 0.102}$ \\
\bottomrule
\end{tabular}
}

\vspace{0.3cm}

\resizebox{\textwidth}{!}{%
\begin{tabular}{lccccccccc}
\toprule
\multicolumn{1}{c}{\multirow{2}{*}{Model}} & \multicolumn{3}{c}{stairs forward}                  & \multicolumn{3}{c}{stairs backward}                 & \multicolumn{3}{c}{obstacle}                        \\
\multicolumn{1}{c}{}                       & mean            & mean@25         & mean@50         & mean            & mean@25         & mean@50         & mean            & mean@25         & mean@50         \\
\midrule
Our                                        & $\mathbf{0.826 \pm 0.024}$ & $\mathbf{0.743 \pm 0.021}$ & $\mathbf{0.775 \pm 0.024}$ & $\mathbf{0.791 \pm 0.039}$ & $\mathbf{0.710 \pm 0.038}$ & $\mathbf{0.744 \pm 0.039}$ & $\mathbf{0.882 \pm 0.004}$ & $\mathbf{0.799 \pm 0.004}$ & $\mathbf{0.835 \pm 0.004}$ \\
CTS                                        & $0.799 \pm 0.051$          & $0.722 \pm 0.050$          & $0.746 \pm 0.050$          & $0.666 \pm 0.056$          & $0.587 \pm 0.053$          & $0.617 \pm 0.055$          & $0.569 \pm 0.008$          & $0.483 \pm 0.007$          & $0.521 \pm 0.008$          \\
HIM                                        & $0.560 \pm 0.060$          & $0.486 \pm 0.051$          & $0.514 \pm 0.054$          & $0.735 \pm 0.082$          & $0.663 \pm 0.068$          & $0.693 \pm 0.074$          & $0.460 \pm 0.028$          & $0.385 \pm 0.020$          & $0.418 \pm 0.024$          \\
DreamWaQ                                   & $0.588 \pm 0.086$          & $0.515 \pm 0.075$          & $0.542 \pm 0.079$          & $0.680 \pm 0.056$          & $0.602 \pm 0.054$          & $0.633 \pm 0.055$          & $0.355 \pm 0.020$          & $0.280 \pm 0.015$          & $0.315 \pm 0.017$          \\
\bottomrule
\end{tabular}
}
\end{table*}

\begin{figure}[htbp]
    \centering
    \includegraphics[width=\linewidth]{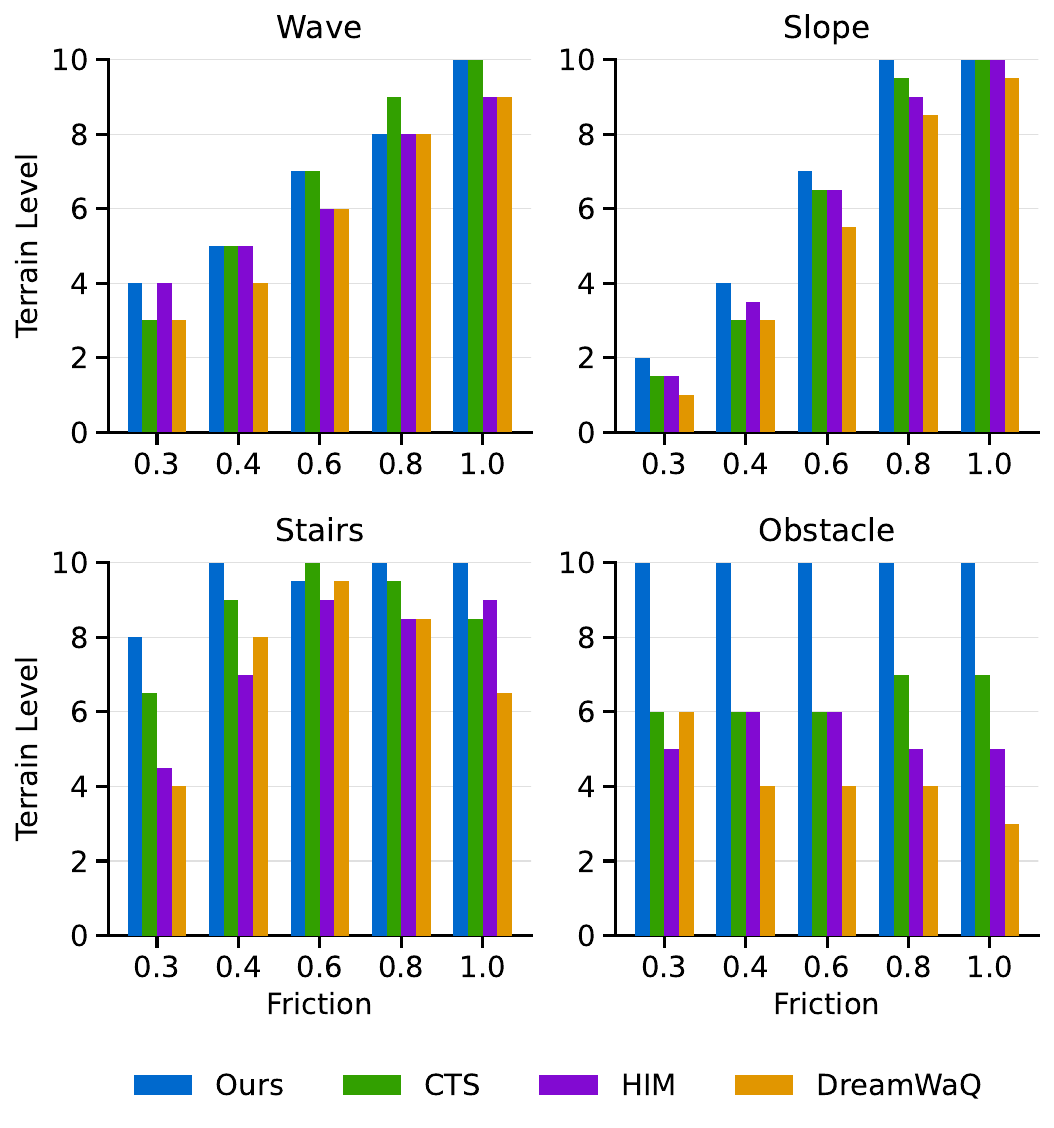}
    \caption{Maximum terrain difficulty levels achieved by various models under a subset of friction coefficients (ranging from 0.2 to 1.0 in increments of 0.1).}
    \label{fig:level_vs_friction_full}
\end{figure}

\begin{figure}[!h]
    \centering
    \begin{subfigure}{0.48\linewidth}
        \centering
        \includegraphics[width=\linewidth]{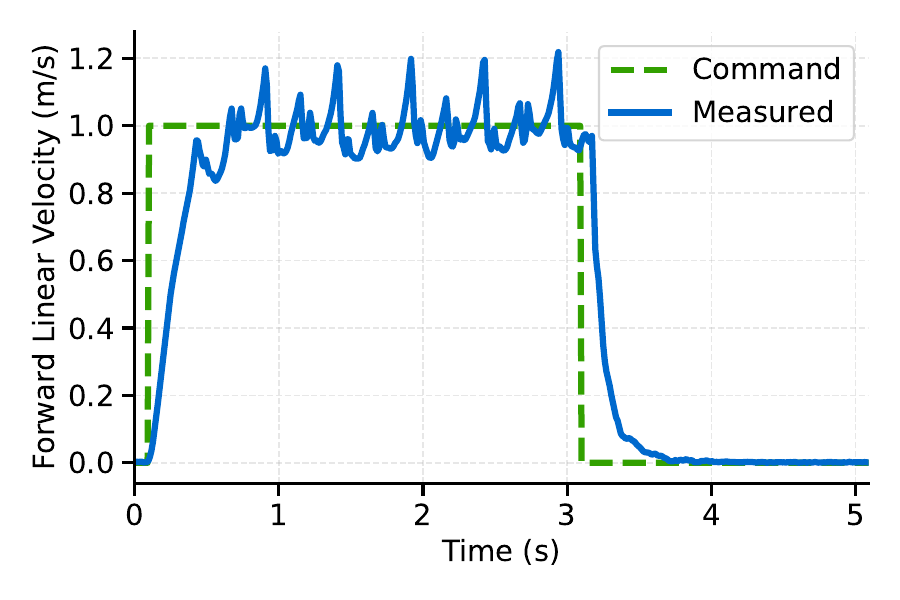}
        \caption{With $v_x^{\text{cmd}}=1.0$ m/s}
        \label{fig:x_vel}
    \end{subfigure}
    \hfill
    \begin{subfigure}{0.48\linewidth}
        \centering
        \includegraphics[width=\linewidth]{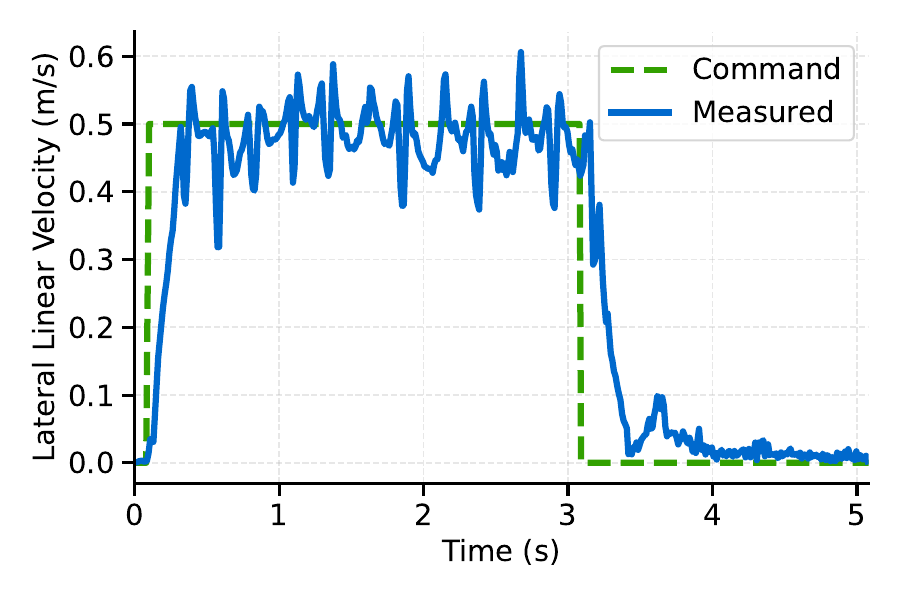}
        \caption{With $v_y^{\text{cmd}}=0.5$ m/s}
        \label{fig:y_vel}
    \end{subfigure}

    \vspace{2mm}

    \begin{subfigure}{0.48\linewidth}
        \centering
        \includegraphics[width=\linewidth]{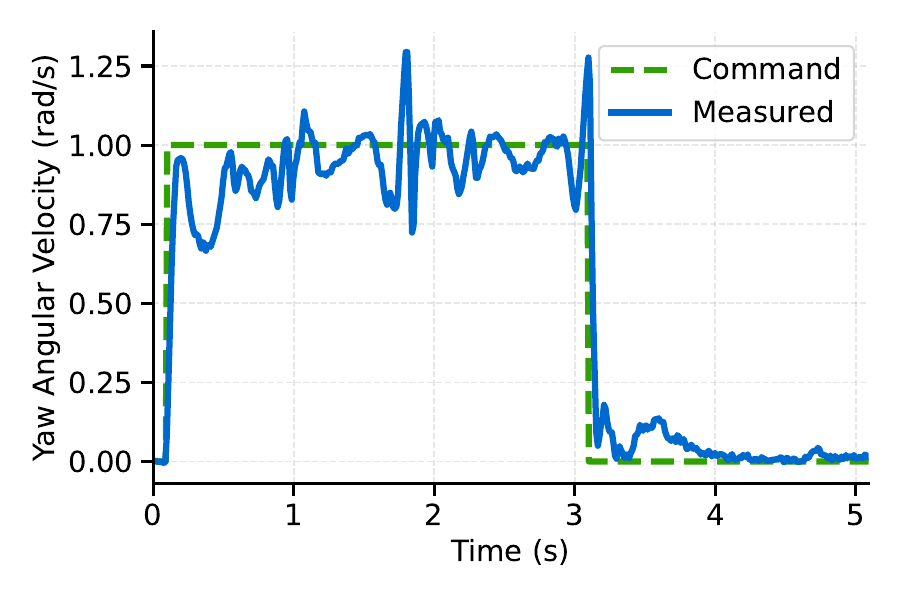}
        \caption{With $\omega_z^{\text{cmd}}=1.0$ rad/s}
        \label{fig:z_vel}
    \end{subfigure}
    \hfill
    \begin{subfigure}{0.48\linewidth}
        \centering
        \includegraphics[width=\linewidth]{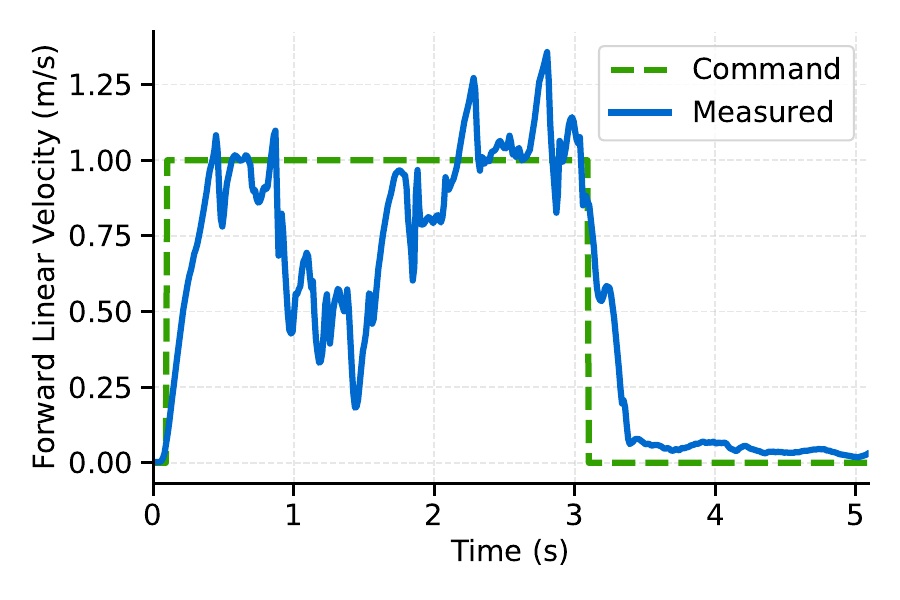}
        \caption{\footnotesize 10 cm stairs with $v_x = 1.0 \text{ m/s}$}
        \label{fig:stairs}
    \end{subfigure}

    \caption{The green dashed lines represent the ground-truth velocities measured by the motion capture system at a sampling frequency of 90 Hz, while the blue solid lines denote the corresponding target command values automatically transmitted to the Unitree Go2 via a pre-defined evaluation program.}
    \label{fig:mocap_metrics_comparison}
\end{figure}

\end{document}